\documentclass{article}

\PassOptionsToPackage{numbers, compress}{natbib}

\usepackage[final]{neurips_2019}

\usepackage[utf8]{inputenc} 
\usepackage[T1]{fontenc}    
\usepackage{hyperref}       
\usepackage{url}            
\usepackage{booktabs}       
\usepackage{amsfonts}       
\usepackage{nicefrac}       
\usepackage{microtype}      
\usepackage{graphicx}
\usepackage{subcaption}
\usepackage{amsmath}
\usepackage{float}
\usepackage{booktabs,multirow}
\usepackage{caption}
\usepackage{subcaption}
\usepackage{color}
\usepackage{url}
\usepackage{multirow}
\usepackage[]{todonotes}
\usepackage[ruled, vlined, linesnumbered]{algorithm2e}
\usepackage{algorithmic}
\usepackage{rotating}
\usepackage{adjustbox}

\newcommand{\loss}{\ell}

\newcommand{\vparam}{\vw}
\newcommand{\param}{w}
\newcommand{\minibatch}{\mathcal{M}}

\newcommand{\comment}[1]{}

\newcommand{\dkls}[3]{\mathbb{D}_{KL}^{#1}[#2 \, \|\, #3]}

\newcommand\cut[1]{}

\newcommand{\squishlist}{
   \begin{list}{$\bullet$}
    { \setlength{\itemsep}{0pt}      \setlength{\parsep}{3pt}
      \setlength{\topsep}{3pt}       \setlength{\partopsep}{0pt}
      \setlength{\leftmargin}{1.5em} \setlength{\labelwidth}{1em}
      \setlength{\labelsep}{0.5em} } }

\newcommand{\squishlisttwo}{
   \begin{list}{$\bullet$}
    { \setlength{\itemsep}{0pt}    \setlength{\parsep}{0pt}
      \setlength{\topsep}{0pt}     \setlength{\partopsep}{0pt}
      \setlength{\leftmargin}{2em} \setlength{\labelwidth}{1.5em}
      \setlength{\labelsep}{0.5em} } }

\newcommand{\squishend}{
    \end{list}  }

{}
{}
{}
{}

{}

{}

\newcommand{\half}{\mbox{$\frac{1}{2}$}}

\newcommand{\real}{\mbox{$\mathbb{R}$}}

\newcommand{\rnd}[1]{\left(#1\right)}
\newcommand{\sqr}[1]{\left[#1\right]}

\newcommand{\myexpect}{\mathbb{E}}

\newcommand{\gauss}{\mbox{${\cal N}$}}

\newcommand{\myvec}[1]{\mbox{$\mathbf{#1}$}}
\newcommand{\myvecsym}[1]{\mbox{$\boldsymbol{#1}$}}

\newcommand{\vbeta}{\mbox{$\myvecsym{\beta}$}}

\newcommand{\vgamma}{\mbox{$\myvecsym{\gamma}$}}

\newcommand{\vmu}{\mbox{$\myvecsym{\mu}$}}

\newcommand{\vlambda}{\mbox{$\myvecsym{\lambda}$}}

\newcommand{\vsigma}{\mbox{$\myvecsym{\sigma}$}}
\newcommand{\vSigma}{\mbox{$\myvecsym{\Sigma}$}}

\newcommand{\va}{\mbox{$\myvec{a}$}}

\newcommand{\vf}{\mbox{$\myvec{f}$}}
\newcommand{\vg}{\mbox{$\myvec{g}$}}
\newcommand{\vh}{\mbox{$\myvec{h}$}}

\newcommand{\vm}{\mbox{$\myvec{m}$}}

\newcommand{\vp}{\mbox{$\myvec{p}$}}

\newcommand{\vs}{\mbox{$\myvec{s}$}}

\newcommand{\vw}{\mbox{$\myvec{w}$}}

\newcommand{\vx}{\mbox{$\myvec{x}$}}

\newcommand{\vy}{\mbox{$\myvec{y}$}}

\newcommand{\vA}{\mbox{$\myvec{A}$}}

\newcommand{\vI}{\mbox{$\myvec{I}$}}

\newcommand{\vM}{\mbox{$\myvec{M}$}}

\newcommand{\vS}{\mbox{$\myvec{S}$}}

\newcommand{\vW}{\mbox{$\myvec{W}$}}

\newcommand{\diag}{\mbox{$\mbox{diag}$}}

\newcommand{\calD}{\mbox{${\cal D}$}}

\newcommand{\data}{\calD}

\newcommand{\be}{\begin{equation}}
\newcommand{\ee}{\end{equation}}
\newcommand{\bea}{\begin{eqnarray}}
\newcommand{\eea}{\end{eqnarray}}
\newcommand{\beaa}{\begin{eqnarray*}}
\newcommand{\eeaa}{\end{eqnarray*}}

\renewcommand{\vec}[1]{\mathbf{#1}}
\newcommand{\matr}[1]{\mathbf{#1}}

\title{Practical Deep Learning with Bayesian Principles}

 \author{
Kazuki Osawa,$^1$ Siddharth Swaroop,$^{2,*}$ Anirudh Jain,$^{3,*,\dagger}$ Runa Eschenhagen,$^{4,\dagger}$ \\
\bf{Richard E. Turner,$^{2}$ Rio Yokota,$^1$ Mohammad Emtiyaz Khan$^{5,\ddagger}$}. \\\\
$^1$ Tokyo Institute of Technology, Tokyo, Japan\\
$^2$ University of Cambridge, Cambridge, UK \\
$^3$ Indian Institute of Technology (ISM), Dhanbad, India\\
$^4$ University of Osnabrück, Osnabrück, Germany\\
$^5$ RIKEN Center for AI Project, Tokyo, Japan
}

\begin{document}

\maketitle
\let\svthefootnote\thefootnote
\let\thefootnote\relax\footnotetext{* These two authors contributed equally.}
\let\thefootnote\relax\footnotetext{$\dagger$ This work is conducted during an internship at RIKEN Center for AI project.}
\let\thefootnote\relax\footnotetext{$\ddagger$ Corresponding author: \tt{emtiyaz.khan@riken.jp}}
\addtocounter{footnote}{0}\let\thefootnote\svthefootnote

\begin{abstract}
    Bayesian methods promise to fix many shortcomings of deep learning, but they are impractical and rarely match the performance of standard methods, let alone improve them.
In this paper, we demonstrate practical training of deep networks with natural-gradient variational inference.
By applying techniques such as batch normalisation, data augmentation, and distributed training, we achieve similar performance in about the same number of epochs as the Adam optimiser, even on large datasets such as ImageNet. 
Importantly, the benefits of Bayesian principles are preserved: predictive probabilities are well-calibrated, uncertainties on out-of-distribution data are improved, and continual-learning performance is boosted. This work enables practical deep learning while preserving benefits of Bayesian principles. A PyTorch implementation\footnote{ The code is available at
\href{https://github.com/team-approx-bayes/dl-with-bayes}{\texttt{https://github.com/team-approx-bayes/dl-with-bayes.}}} is available as a plug-and-play optimiser.

\end{abstract}

\section{Introduction}
\label{sec:introduction}
Deep learning has been extremely successful in many fields such as computer vision \citep{krizhevsky2012imagenet}, speech processing \citep{hinton2012deep}, and natural-language processing \citep{mikolov2013efficient}, but it is also plagued with several issues that make its application difficult in many other fields.
For example, it requires a large amount of high-quality data and it can overfit when dataset size is small.
Similarly, sequential learning can cause forgetting of past knowledge \citep{kirkpatrick2017overcoming}, and lack of reliable confidence estimates and other robustness issues can make it vulnerable to adversarial attacks \citep{bradshaw2017adversarial}.
Ultimately, due to such issues, application of deep learning remains challenging, especially for applications where human lives are at risk.

Bayesian principles have the potential to address such issues. For example, we can represent uncertainty using the posterior distribution, enable sequential learning using Bayes' rule, and reduce overfitting with Bayesian model averaging \cite{hoeting1999bayesian}.
The use of such Bayesian principles for neural networks has been advocated from very early on.
Bayesian inference on neural networks were all proposed in the 90s, e.g., by using MCMC methods \citep{neal95}, Laplace's method \citep{mackay1991thesis}, and variational inference (VI) \citep{hinton1993keeping,barber1998ensemble,saul1996mean, anderson1987mean}.
Benefits of Bayesian principles are even discussed in machine-learning textbooks \citep{mackay2003information,bishop2006pattern}.
Despite this, they are rarely employed in practice.
This is mainly due to computational concerns, unfortunately overshadowing their theoretical advantages.

The difficulty lies in the computation of the posterior distribution, which is especially challenging for deep learning.
Even approximation methods, such as VI and MCMC, have historically been difficult to scale to large datasets such as ImageNet \citep{russakovsky2015imagenet}.
Due to this, it is common to use less principled approximations, such as MC-dropout \citep{yarin16dropout}, even though they are not ideal when it comes to fixing the issues of deep learning.
For example, MC-dropout is unsuitable for continual learning \cite{kirkpatrick2017overcoming} since its posterior approximation does not have mass over the whole weight space. 
It is also found to perform poorly for sequential decision making \citep{riquelme2018deep}.
The form of the approximation used by such methods is usually rigid and cannot be easily improved, e.g., to other forms such as a mixture of Gaussians.
The goal of this paper is to make more principled Bayesian methods, such as VI, practical for deep learning, thereby helping researchers tackle its key limitations.

\begin{figure}[!t]
    \centering
    \includegraphics[width=\textwidth]
    {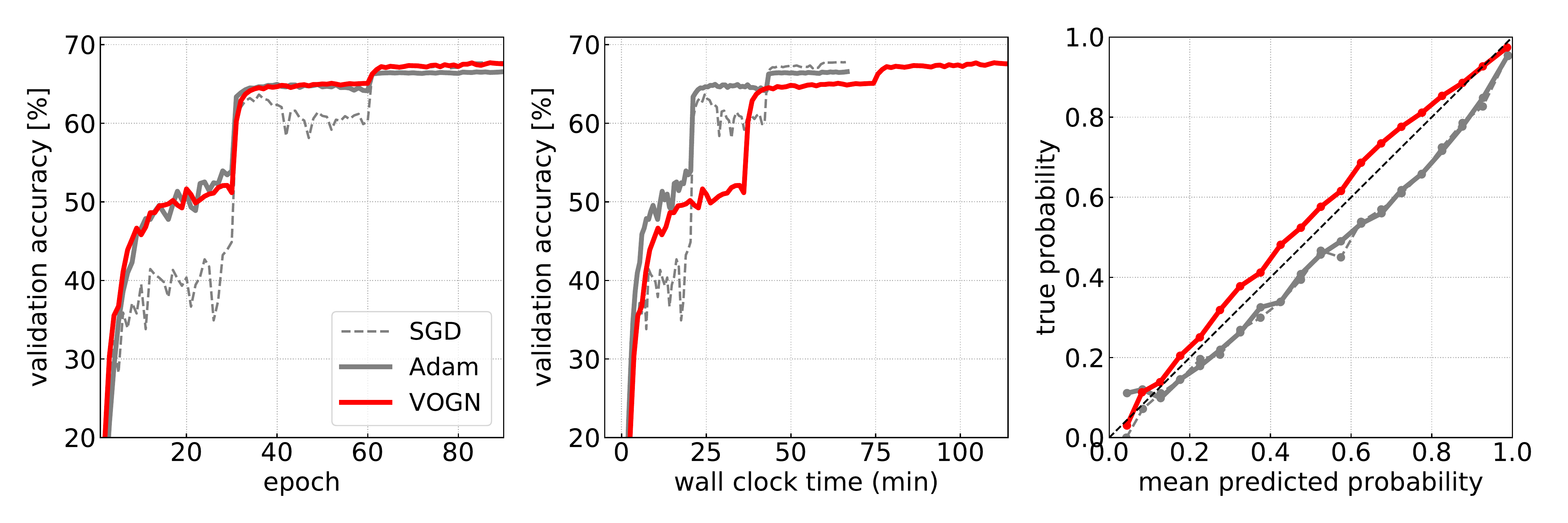}
    \caption{Comparing VOGN \citep{khan2018fast}, a natural-gradient VI method, to Adam and SGD, training ResNet-18 on ImageNet. The two left plots show that VOGN and Adam have similar convergence behaviour and achieve similar performance in about the same number of epochs. 
    VOGN achieves 67.38\% on validation compared to 66.39\% by Adam and 67.79\% by SGD. Run-time of VOGN is 76 seconds per epoch compared to 44 seconds for Adam and SGD. The rightmost figure shows the calibration curve. VOGN gives calibrated predictive probabilities (the diagonal represents perfect calibration).}
    \label{fig:resnet18_imagenet_new}
\end{figure}

We demonstrate practical training of deep networks by using recently proposed natural-gradient VI methods. 
These methods resemble the Adam optimiser, enabling us to leverage existing techniques for initialisation, momentum, batch normalisation, data augmentation, and distributed training. As a result, we obtain similar performance in about the same number of epochs as Adam when training many popular deep networks (e.g., LeNet, AlexNet, ResNet) on datasets such as CIFAR-10 and ImageNet (see Fig.~\ref{fig:resnet18_imagenet_new}).
The results show that, despite using an approximate posterior, the training methods preserve the benefits coming from Bayesian principles.
Compared to standard deep-learning methods, the predictive probabilities are well-calibrated, uncertainties on out-of-distribution inputs are improved, and performance for continual-learning tasks is boosted.
Our work shows that practical deep learning is possible with Bayesian methods and aims to support further research in this area.

\textbf{Related work.} Previous VI methods, notably by \citet{graves2011practical} and \citet{blundell2015weight}, require significant implementation and tuning effort to perform well, e.g., on convolution neural networks (CNN).
Slow convergence is found to be especially problematic for sequential problems \citep{riquelme2018deep}.
There appears to be no reported results with complex networks on large problems, such as ImageNet. 
Our work solves these issues by applying deep-learning techniques to natural-gradient VI \citep{khan2018fast, zhang2018noisy}.

In their paper, \citet{zhang2018noisy} also employed data augmentation and batch normalisation for a natural-gradient method called Noisy K-FAC (see Appendix \ref{app:noisykfac}) and showed results on VGG on CIFAR-10.
However, a mean-field method called Noisy Adam was found to be unstable with batch normalisation. 
In contrast, we show that a similar method, called Variational Online Gauss-Newton (VOGN), proposed by \citet{khan2018fast}, works well with such techniques.
We show results for distributed training with Noisy K-FAC on Imagenet, but do not provide extensive comparisons since tuning it is time-consuming.
Many of our techniques can speed-up Noisy K-FAC, which is promising.

Many other approaches have recently been proposed to compute posterior approximations by training deterministic networks \citep{ ritter2018scalable, maddox2019simple, mandt2017stochastic}.
Similarly to MC-dropout, their posterior approximations are not flexible, making it difficult to improve the accuracy of their approximations.
On the other hand, VI offers a much more flexible alternative to apply Bayesian principles to deep learning.

\section{Deep Learning with Bayesian Principles and Its Challenges}
\label{sec:theory}
The success of deep learning is partly due to the availability of scalable and practical methods for training deep neural networks (DNNs).
Network training is formulated as an optimisation problem where a loss between the data and the DNN's predictions is minimised. For example, in a supervised learning task with a dataset $\data$ of $N$ inputs $\vx_i$ and corresponding outputs $\vy_i$ of length $K$, we minimise a loss of the following form: $\bar{\loss}(\vparam) + \delta\vw^\top\vw$, where $\bar{\loss}(\vparam) := \frac{1}{N} \sum_i \loss( \vy_i, \vf_\param(\vx_i) )$, $\vf_\param (\vx) \in \real^K$ denotes the DNN outputs with weights $\vparam$, $\loss(\vy,\vf)$ denotes a differentiable loss function between an output $\vy$ and the function $\vf$, and $\delta>0$ is the $L_2$ regulariser.\footnote{This regulariser is sometimes set to 0 or a very small value.}
Deep learning relies on stochastic-gradient (SG) methods to minimise such loss functions. The most commonly used optimisers, such as stochastic-gradient descent (SGD), RMSprop \citep{hintonTieleman}, and Adam \citep{kingma2014adam}, take the following form\footnote{Alternate versions with weight-decay and momentum differ from this update \citep{loshchilov2018decoupled}. We present a form useful to establish the connection between SG methods and natural-gradient VI.} (all operations below are element-wise):
\begin{align}
    \vparam_{t+1} \leftarrow \vparam_t - \alpha_t \frac{\hat{\vg}(\vparam_t) + \delta\vparam_t}{{\sqrt{\vs_{t+1}} + \epsilon}} ,
    \quad\quad \vs_{t+1} \leftarrow (1-\beta_t) \vs_t + \beta_t \rnd{\hat{\vg} (\vparam_t) + \delta\vparam_t }^2,
    \label{eq:sg_updates}
\end{align}
where $t$ is the iteration, $\alpha_t>0$ and $0<\beta_t<1$ are learning rates, $\epsilon>0$ is a small scalar, and $\hat{\vg}(\vparam)$ is the stochastic gradients at $\vparam$ defined as follows: $\hat{\vg}(\vparam) := \frac{1}{M} \sum_{i\in\minibatch_t} \nabla_\param \loss(\vy_i, \vf_{\param}(\vx_i))$ using a minibatch $\minibatch_t$ of $M$ data examples.
This simple update scales extremely well and can be applied to very large problems. With techniques such as initialisation protocols, momentum, weight-decay, batch normalisation, and data augmentation, it also achieves good performance for many problems.

In contrast, the full Bayesian approach to deep learning is computationally very expensive.
The posterior distribution can be obtained using Bayes' rule: 
$p(\vparam|\data) = \exp\rnd{-N\bar{\loss}(\vparam)/\tau} p(\vparam)/p(\data)$ where $0<\tau\le1$.\footnote{This is a tempered posterior \citep{Vovk:1990:AS:92571.92672} setup where $\tau$ is set $\ne 1$ when we expect model misspecification and/or adversarial examples \citep{ghosal2017fundamentals}. Setting $\tau=1$ recovers standard Bayesian inference.}
This is costly due to the computation of the marginal likelihood $p(\data)$,  a high-dimensional integral that is difficult to compute for large networks.
Variational inference (VI) is a principled approach to more scalably estimate an approximation to $p(\vparam | \data)$. The main idea is to employ a parametric approximation, e.g., a Gaussian $q(\vparam) := \gauss(\vparam|\vmu,\vSigma)$ with mean $\vmu$ and covariance $\vSigma$.
The parameters $\vmu$ and $\vSigma$ can then be estimated by maximising the \emph{evidence lower bound (ELBO)}:
   \begin{align}
	\textrm{ELBO: } \quad  \mathcal{L}(\vmu,\vSigma) := -N \myexpect_q \sqr{\bar{\loss}(\vparam)} - \tau\dkls{}{q(\vparam)}{p(\vparam)} , \label{eq:elbo}
	\end{align}
where $\mathbb{D}_{KL}[\cdot]$ denotes the Kullback-Leibler divergence.
By using more complex approximations, we can further reduce the approximation error, but at a computational cost.
By formulating Bayesian inference as an optimisation problem, VI enables a practical application of Bayesian principles.

Despite this, VI has remained impractical for training large deep networks on large datasets. 
Existing methods, such as \citet{graves2011practical} and \citet{blundell2015weight}, directly apply popular SG methods to optimise the variational parameters in the ELBO, yet they fail to get a reasonable performance on large problems, usually converging very slowly.
The failure of such direct applications of deep-learning methods to VI is not surprising. The techniques used in one field may not directly lead to improvements in the other, but it will be useful if they do, e.g., if we can optimise the ELBO in a way that allows us to exploit the tricks and techniques of deep learning and boost the performance of VI.
The goal of this work is to do just that.
We now describe our methods in detail.

\section{Practical Deep Learning with Natural-Gradient Variational Inference}
\label{sec:methods}
In this paper, we propose natural-gradient VI methods for practical deep learning with Bayesian principles.
The natural-gradient update takes a simple form when estimating exponential-family approximations \citep{khan2018fast1, khan2017conjugate}. When $p(\vparam) := \gauss(\vparam|0,\vI/\delta)$, the update of the natural-parameter $\vlambda$ is performed by using the stochastic gradient of the \emph{expected regularised-loss}:
\begin{align}
    \vlambda_{t+1} = (1-\tau\rho) \vlambda_t - \rho \nabla_\mu \myexpect_q \sqr{\bar{\loss}(\vparam) + \half \tau\delta \vparam^\top \vparam},
    \label{eq:ngvi}
\end{align}
where $\rho>0$ is the learning rate, and we note that the stochastic gradients are computed with respect to $\vmu$, the \emph{expectation parameters} of $q$.
The \emph{moving average} above helps to deal with the stochasticity of the gradient estimates, and is very similar to the moving average used in deep learning (see \eqref{eq:sg_updates}).
When $\tau$ is set to 0, the update essentially minimises the regularised loss (see Section 5 in \citet{khan2018fast}).
These properties of natural-gradient VI makes it an ideal candidate for deep learning. 

Recent work by \citet{khan2018fast} and \citet{zhang2018noisy} further show that, when $q$ is Gaussian, the update \eqref{eq:ngvi} assumes a form that is strikingly similar to the update \eqref{eq:sg_updates}.
For example, the Variational Online Gauss-Newton (VOGN) method of \citet{khan2018fast} estimates a Gaussian with mean $\vmu_t$ and a diagonal covariance matrix $\vSigma_t$ using the following update:
\begin{align}
    \vmu_{t+1} &\leftarrow \vmu_t - \alpha_t \frac{\hat{\vg}(\vparam_t) + \tilde{\delta}\vmu_t}{\vs_{t+1} + \tilde{\delta}} , \quad
    \vs_{t+1} \leftarrow (1- \tau\beta_t) \vs_{t}+ \beta_t \frac{1}{M} \sum_{i\in\minibatch_t} \rnd{\vg_i(\vparam_t)}^2,
    \label{eq:vogn}
\end{align}
where $\vg_i(\vparam_t) := \nabla_\param \loss(y_i, f_{\param_t}(\vx_i))$, $\vparam_t \sim\gauss(\vw|\vmu_t,\vSigma_t)$ with $\vSigma_t := \diag(1/(N(\vs_t + \tilde{\delta})))$, $\tilde{\delta} := \tau\delta/N$, and $\alpha_t, \beta_t>0$ are learning rates.
Operations are performed element-wise.
Similarly to \eqref{eq:sg_updates}, the vector $\vs_t$ adapts the learning rate and is updated using a moving average.

A major difference in VOGN is that the update of $\vs_t$ is now based on a Gauss-Newton approximation \cite{graves2011practical} which uses $\frac{1}{M}\sum_{i\in\minibatch_t} (\vg_i(\vparam_t))^2$.
This is fundamentally different from the SG update in \eqref{eq:sg_updates} which instead uses the gradient-magnitude $(\frac{1}{M}\sum_{i\in\minibatch_t} \vg_i(\vparam_t) + \delta\vw_t)^2$ \citep{bottou2016optimization}.  
The first approach uses the sum \emph{outside} the square while the second approach uses it \emph{inside}.
VOGN is therefore a second-order method and, similarly to Newton's method, does not need a square-root over $\vs_t$. 
Implementation of this step requires an additional calculation (see Appendix~\ref{app:GN implementation}) which makes VOGN a bit slower than Adam, but VOGN is expected to give better variance estimates (see Theorem 1 in \citet{khan2018fast}).

\begin{figure}[!t]
    \begin{minipage}[t]{0.60\textwidth}
    \begin{algorithm}[H]
      \footnotesize
      \caption{Variational Online Gauss Newton (VOGN)}
      \label{alg:vogn}
            \begin{algorithmic}[1]
            \STATE Initialise $\vmu_0$, $\vs_0$, $\vm_0$. \label{alg:initialise}
            \STATE $N\leftarrow \rho N$, $\tilde{\delta} \leftarrow \tau \delta/N$. \label{alg:da}
                \REPEAT
                    \STATE Sample a minibatch $\minibatch$ of size $M$. 
                    \STATE Split $\minibatch$ into each GPU  (local minibatch $\minibatch_{local}$).
                    \FOR{ each GPU in parallel}
                    \FOR{$k=1,2,\ldots,K$ } 
                        \STATE Sample $\epsilon \sim \gauss (\mathbf{0}, \vI)$.
                        \STATE $\vparam^{(k)} \leftarrow \vmu + \epsilon \vsigma$ with $\vsigma \leftarrow (1/ (N (\vs + \tilde{\delta}+\gamma)))^{1/2}$.
                        \STATE Compute $\vg_i^{(k)} \leftarrow \nabla_\param \loss(\vy_i,\vf_{\param^{(k)}}(\vx_i)), \forall i\in\minibatch_{local}$ using the method described in Appendix \ref{app:GN implementation}.
                        \STATE $\hat{\vg}_k \leftarrow \frac{1}{M} \sum_{i\in\minibatch_{local}} \vg_i^{(k)}$.
                        \STATE $\hat{\vh}_k \leftarrow \frac{1}{M} \sum_{i\in\minibatch_{local}} (\vg_i^{(k)})^2 $ \label{alg:gn_step}.
                    \ENDFOR
                    \STATE $\hat{\vg} \leftarrow \frac{1}{K} \sum_{k=1}^{K} \hat\vg_k$ and $\hat{\vh} \leftarrow \frac{1}{K} \sum_{k=1}^{K} \hat{\vh}_k$.
                    \ENDFOR
                    \STATE AllReduce $\hat{\vg},\hat{\vh}$.
                    \STATE $\vm \leftarrow \beta_1 \vm + (\hat{\vg} + \tilde{\delta} \vmu)$. \label{alg:momentum}
                    \STATE $\vs \leftarrow (1 - \tau\beta_2) \vs + \beta_2 \hat\vh$.
                    \STATE $\vmu \leftarrow \vmu - \alpha {\vm}/(\vs + \tilde{\delta}+\gamma)$.
                \UNTIL{stopping criterion is met}
            \end{algorithmic}
    \end{algorithm}
    \end{minipage}
    \hfill
    \begin{minipage}[t]{0.4\textwidth}
        \begin{figure}[H]
        \centering
        \includegraphics[height=1.92in]{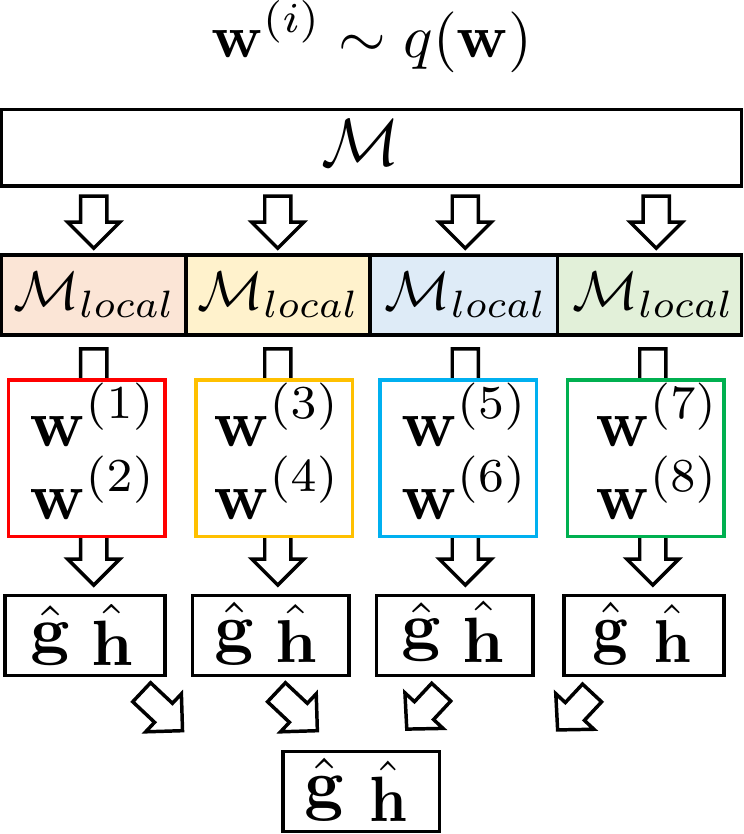}
        \end{figure}
        \small
        \centering
        \begin{tabular}{l c}
            \toprule
            Learning rate & $\alpha$ \\
            Momentum rate & $\beta_1$ \\
            Exp. moving average rate & $\beta_2$ \\
            Prior precision & $\delta$ \\
            External damping factor & $\gamma$ \\
            Tempering parameter & $\tau$ \\
            \# MC samples for training & $K$ \\
            Data augmentation factor & $\rho$ \\
            \bottomrule
        \end{tabular}
        \label{tab:training_settings}
    \end{minipage}
    \caption{A pseudo-code for our distributed VOGN algorithm is shown in Algorithm \ref{alg:vogn}, 
    and the distributed scheme is shown in the right figure. 
    The computation in line 10 requires an extra calculation (see Appendix~\ref{app:GN implementation}), making VOGN slower than Adam. The bottom table gives a list of algorithmic hyperparameters needed for VOGN.
    }
    \label{fig:distributed_vi}
\end{figure}

The main contribution of this paper is to demonstrate practical training of deep networks using VOGN. 
Since VOGN takes a similar form to SG methods, we can easily borrow existing deep-learning techniques to improve performance.
We will now describe these techniques in detail.
Pseudo-code for VOGN is shown in Algorithm \ref{alg:vogn}.

\textbf{Batch normalisation:} Batch normalisation \citep{ioffe2015batchnorm} has been found to significantly speed up and stabilise training of neural networks, and is widely used in  deep learning.
BatchNorm layers are inserted between neural network layers.  They help stabilise each layer's input distribution by normalising the running average of the inputs' mean and variance.
In our VOGN implementation, we simply use the existing implementation with default hyperparameter settings.
We do not apply L2 regularisation and weight decay to BatchNorm parameters, like in \citet{goyal2017accurate}, or maintain uncertainty over the BatchNorm parameters.
This straightforward application of batch normalisation works for VOGN.

\textbf{Data Augmentation:} When training on image datasets, data augmentation (DA) techniques can improve performance drastically \cite{goyal2017accurate}.
We consider two common real-time data augmentation techniques: random cropping and horizontal flipping.
After randomly selecting a minibatch at each iteration, we use a randomly selected cropped version of all images.
Each image in the minibatch has a $50\%$ chance of being horizontally flipped.

We find that directly applying DA gives slightly worse performance than expected, and also affects the calibration of the resulting uncertainty. However, DA increases the effective sample size. We therefore modify it to be $\rho N$ where $\rho\ge 1$, improving performance (see step \ref{alg:da} in Algorithm \ref{alg:vogn}).
The reason for this performance boost might be due to the complex relationship between the regularisation $\delta$ and $N$.
For the regularised loss $\bar{\loss}(\vparam) + \delta \vparam^\top \vparam$, the two are unidentifiable, i.e., we can multiply $\delta$ by a constant and reduce $N$ by the same constant without changing the minimum.
However, in a Bayesian setting (like in \eqref{eq:elbo}), the two quantities are separate, and therefore changing the data might also change the optimal prior variance hyperparameter in a complicated way.
This needs further theoretical investigations, but our simple fix of scaling $N$ seems to work well in the experiments.

We set $\rho$ by considering the specific DA techniques used. When training on CIFAR-10, the random cropping DA step involves first padding the 32x32 images to become of size 40x40, and then taking randomly selected 28x28 cropped images. We consider this as effectively increasing the dataset size by a factor of 5 (4 images for each corner, and one central image). The horizontal flipping DA step doubles the dataset size (one dataset of unflipped images, one for flipped images). Combined, this gives $\rho=10$.
Similar arguments for ImageNet DA techniques give $\rho=5$. Even though $\rho$ is another hyperparameter to set, we find that its precise value does not matter much. Typically, after setting an estimate for $\rho$, tuning $\delta$ a little seems to work well (see Appendix \ref{app:prior variance}).

\textbf{Momentum and initialisation:} It is well known that both momentum and good initialisation can improve the speed of convergence for SG methods in deep learning \cite{sutskever2013importance}.
Since VOGN is similar to Adam, we can implement momentum in a similar way.
This is shown in step \ref{alg:momentum} of Algorithm \ref{alg:vogn}, where $\beta_1$ is the momentum rate.
We initialise the mean $\vmu$ in the same way the weights are initialised in Adam (we use {\tt init.xavier\_normal} in PyTorch \citep{glorot2010understanding}).
For the momentum term $\vm$, we use the same initialisation as Adam (initialised to 0). 
VOGN requires an additional initialisation for the variance $\vsigma^2$. For this, we first run a forward pass through the first minibatch, calculate the average of the squared gradients and initialise the scale $\vs_0$ with it (see step \ref{alg:initialise} in Algorithm \ref{alg:vogn}).
This implies that the variance is initialised to $\vsigma_0^2 = \tau/(N(\vs_0+\tilde{\delta}))$.
For the tempering parameter $\tau$, we use a schedule where it is increased from a small value (e.g., 0.1) to 1.
With these initialisation protocols, VOGN is able to mimic the convergence behaviour of Adam in the beginning.

\textbf{Learning rate scheduling:} A common approach to quickly achieve high validation accuracies is to use a specific learning rate schedule \citep{goyal2017accurate}.
The learning rate (denoted by $\alpha$ in Algorithm \ref{alg:vogn}) is regularly decayed by a factor (typically a factor of 10). The frequency and timings of this decay are usually pre-specified. In VOGN, we use the same schedule used for Adam, which works well.

\textbf{Distributed training:} We also employ distributed training for VOGN to perform large experiments quickly.
We can parallelise computation both over data and Monte-Carlo (MC) samples.
Data parallelism is useful to split up large minibatch sizes. This is followed by averaging over multiple MC samples and their losses on a single GPU.
MC sample parallelism is useful when minibatch size is small, and we can copy the entire minibatch and process it on a single GPU. Algorithm \ref{alg:vogn} and Figure \ref{fig:distributed_vi} illustrate our distributed scheme.
We use a combination of these two parallelism techniques with different MC samples for different inputs.
This theoretically reduces the variance during training (see Equation 5 in \citet{kingma2015variational}), but sometimes requires averaging over multiple MC samples to get a sufficiently low variance in the early iterations. 
Overall, we find that this type of distributed training is essential for fast training on large problems such as ImageNet. 

\textbf{Implementation of the Gauss-Newton update in VOGN:} As discussed earlier, VOGN uses the Gauss-Newton approximation, which is fundamentally different from Adam. 
In this approximation, the gradients on individual data examples are first squared and then averaged afterwards (see step \ref{alg:gn_step} in Algorithm \ref{alg:vogn} which implements the update for $\vs_t$ shown in \eqref{eq:vogn}).
We need extra computation to get access to individual gradients, due to which, VOGN is slower Adam or SGD (e.g., in Fig. \ref{fig:resnet18_imagenet_new}). However, this is not a theoretical limitation and this can be improved if a framework enables an easy computation of the individual gradients.
Details of our implementation are described in Appendix \ref{app:GN implementation}.
This implementation is much more efficient than a naive one where gradients over examples are stored and the sum over the square is computed sequentially. Our implementation usually brings the running time of VOGN to within 2-5 times of the time that Adam takes.

\textbf{Tuning VOGN:} 
Currently, there is no common recipe for tuning the algorithmic hyperparameters for VI, especially for large-scale tasks like ImageNet classification.
One key idea we use in our experiments is to start with Adam hyperparameters and then make sure that VOGN training closely follows an Adam-like trajectory in the beginning of training.
To achieve this, we divide the tuning into an \textit{optimisation part} and a \textit{regularisation part}.
In the \textit{optimisation part}, we first tune the hyperparameters of a deterministic version of VOGN, called the online Gauss-Newton (OGN) method.
This method, described in Appendix~\ref{app:ogn}, is more stable than VOGN since it does not require MC sampling, and can be used as a stepping stone when moving from Adam/SGD to VOGN. 
After reaching a competitive performance to Adam/SGD by OGN, we move to the \textit{regularisation part},
where we tune the prior precision $\delta$, the tempering parameter $\tau$, and the number of MC samples $K$ for VOGN.
We initialise our search by setting the prior precision $\delta$ using the L2-regularisation parameter used for OGN, as well as the dataset size $N$.
Another technique is to warm-up the parameter $\tau$ towards $\tau=1$ (also see the ``momentum and initialisation" part). 
Setting $\tau$ to smaller values usually stabilises the training, and increasing it slowly also helps during tuning.
We also add an \textit{external damping factor} $\gamma>0$ to the moving average $\vs_t$. 
This increases the lower bound of the eigenvalues of the diagonal covariance $\vSigma_t$ and prevents the noise and the step size from becoming too large.
We find that a mix of these techniques works well for the problems we considered.

\section{Experiments}
\label{sec:experiments}

In this section, we present experiments on fitting several deep networks on CIFAR-10 and ImageNet.
Our experiments demonstrate practical training using VOGN on these benchmarks and show performance that is competitive with Adam and SGD. 
We also assess the quality of the posterior approximation, finding that benefits of Bayesian principles are preserved. 

CIFAR-10 \citep{krizhevsky2009cifar} contains 10 classes with 50,000 images for training and 10,000 images for validation. 
For ImageNet, we train with 1.28 million training examples and validate on 50,000 examples, classifying between 1,000 classes.
We used a large minibatch size $M=4,096$ and parallelise them across 128 GPUs (NVIDIA Tesla P100).
We compare the following methods on CIFAR-10: Adam, MC-dropout \cite{yarin16dropout}. For ImageNet, we also compare to SGD, K-FAC, and Noisy K-FAC. We do not consider Noisy K-FAC for other comparisons since tuning is difficult. We compare 3 architectures: LeNet-5, AlexNet, ResNet-18.
We only compare to  Bayes by Backprop (BBB) \cite{blundell2015weight} for CIFAR-10 with LeNet-5 since it is very slow to converge for larger-scale experiments.
We carefully set the hyperparameters of all methods, following the best practice of large distributed training \citep{goyal2017accurate} as the initial point of our hyperparameter tuning.
The full set of hyperparameters is in Appendix \ref{app:vogn hyperparams}.

\subsection{Performance on CIFAR-10 and ImageNet}
\label{sec:matching performance}

\begin{figure}[!t]
    \begin{subfigure}{.58\textwidth}
    \centering
    \includegraphics[height=1.25in]
    {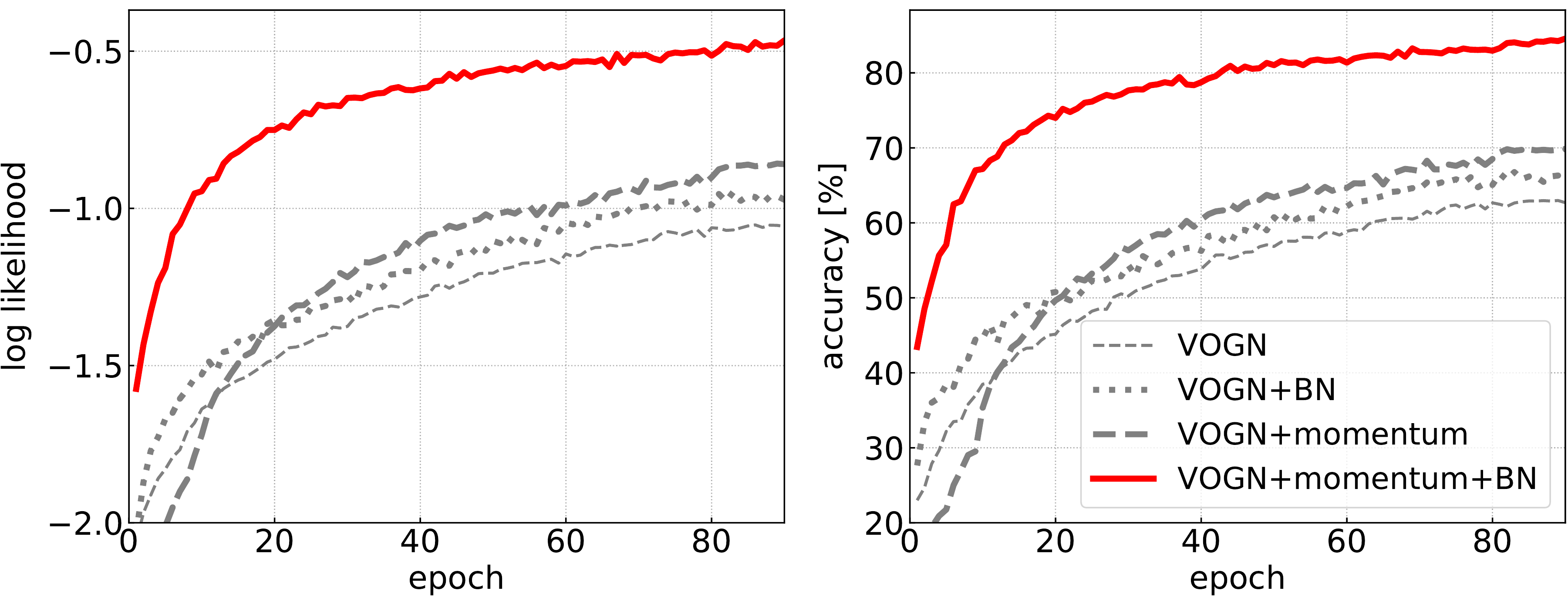}
    \caption{Effect of momentum and batch normalisation.}
    \label{fig:resnet18_cifar10_vogn}
    \end{subfigure}
    \hfill
    \begin{subfigure}{.36\textwidth}
        \centering
        \includegraphics[height=1.25in]
        {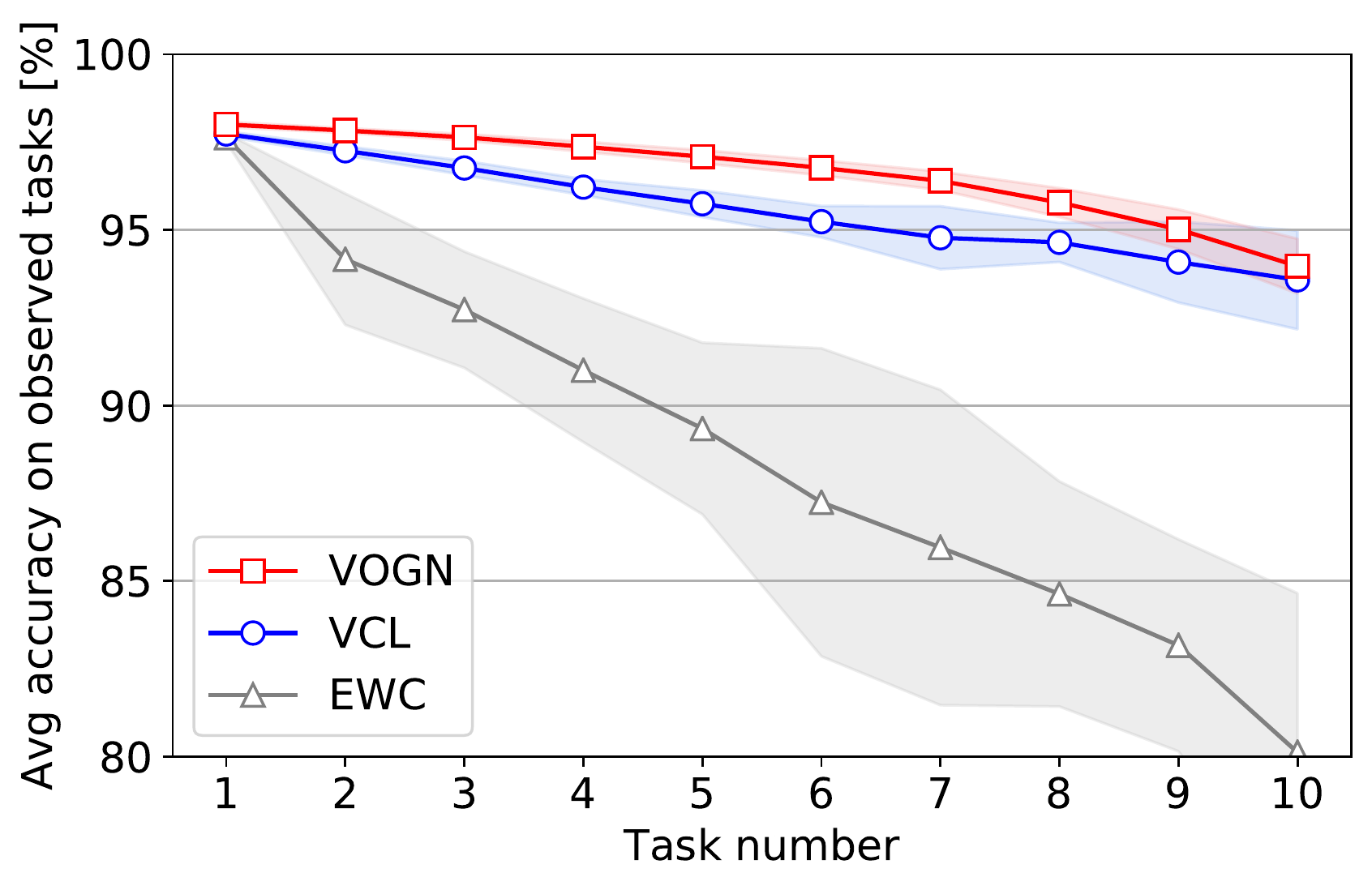}
        \caption{Continual Learning}
        \label{fig:permuted mnist}
    \end{subfigure}
    \caption{ Figure (a) shows that momentum and batch normalisation improve the performance of VOGN. The results are for training ResNet-18 on CIFAR-10. Figure (b) shows comparison for a continual-learning task on the Permuted MNIST dataset. VOGN performs at least as well (average accuracy) as VCL over 10 tasks. We also find that, for each task, VOGN converges much faster, taking only 100 epochs per task as opposed to 800 epochs taken by VCL (plots not shown).}
\end{figure}

We start by showing the effectiveness of momentum and batch normalisation for boosting the performance of VOGN.
Figure \ref{fig:resnet18_cifar10_vogn} shows that these methods significantly speed up convergence and performance (in terms of both accuracy and log likelihoods).

\begin{figure}[t]
    \centering
    \includegraphics[width=\textwidth]
    {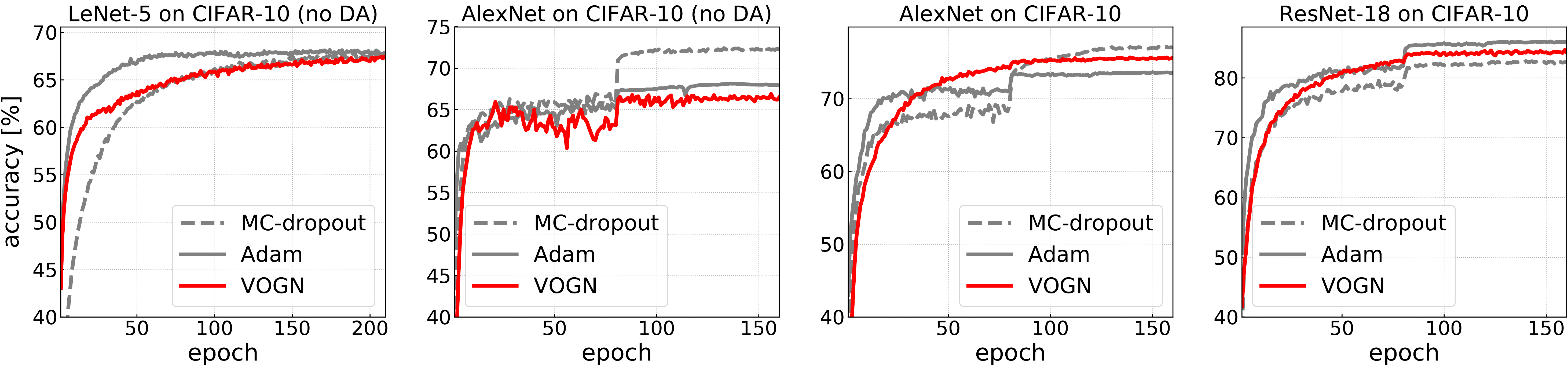}
    \caption{
    Validation accuracy for various architectures trained on CIFAR-10 (DA: Data Augmentation). VOGN's convergence and validation accuracies are comparable to Adam and MC-dropout.
    }
    \label{fig:cifar_many_architectures}
\end{figure}

Figures \ref{fig:resnet18_imagenet_new} and \ref{fig:cifar_many_architectures} compare the convergence of VOGN to Adam (for all experiments), SGD (on ImageNet), and MC-dropout (on the rest). VOGN shows similar convergence and its performance is competitive with these methods.
We also try BBB on LeNet-5, where it converges prohibitively slowly, performing very poorly. We are not able to successfully train other architectures using this approach. 
We found it far simpler to tune VOGN because we can borrow all the techniques used for Adam.
Figure \ref{fig:cifar_many_architectures} also shows the importance of DA in improving performance.

Table \ref{table:full results} gives a final comparison of train/validation accuracies, negative log likelihoods, epochs required for convergence, and run-time per epoch.
We can see that the accuracy, log likelihoods, and the number of epochs are comparable.
VOGN is 2-5 times slower than Adam and SGD. 
This is mainly due to the computation of individual gradients required in VOGN (see the discussion in Section \ref{sec:methods}). 
We clearly see that by using deep-learning techniques on VOGN, we can perform practical deep learning.
This is not possible with methods such as BBB.

Due to the Bayesian nature of VOGN, there are some trade-offs to consider. Reducing the prior precision ($\delta$ in Algorithm \ref{alg:vogn}) results in higher validation accuracy, but also larger train-test gap (more overfitting).
This is shown in Appendix \ref{app:prior variance} for VOGN on ResNet-18 on ImageNet. As expected, when the prior precision is small, performance is similar to non-Bayesian methods.
We also show the effect of changing the effective dataset size $\rho$ in Appendix \ref{app:prior variance}: note that, since we are going to tune the prior variance $\delta$ anyway, it is sufficient to set $\rho$ to its correct order of magnitude.
Another trade-off concerns the number of Monte-Carlo (MC) samples, shown in Appendix \ref{app:monte carlo samples}.
Increasing the number of training MC samples (up to a limit) improves VOGN's convergence rate and stability, but also increases the computation.
Increasing the number of MC samples during testing improves generalisation, as expected due to averaging. 

Finally, a few comments on the performance of the other methods. Adam regularly overfits the training set in most settings, with large train-test differences in both validation accuracy and log likelihood.
One exception is LeNet-5, which is most likely due to the small architecture which results in underfitting (this is consistent with the low validation accuracies obtained).
In contrast to Adam, MC-dropout has small train-test gap, usually smaller than VOGN's.
However, we will see in Section \ref{sec:uncertainty} that this is because of underfitting.
Moreover, the performance of MC-dropout is highly sensitive to the dropout rate (see Appendix \ref{app:MC Dropout sensitivity} for a comparison of different dropout rates).
On ImageNet, Noisy K-FAC performs well too. It is slower than VOGN, but it takes fewer epochs. Overall, wall clock time is about the same as VOGN.

\begin{table}[thbp]
    \centering
    \footnotesize
    \setlength{\tabcolsep}{4pt}
    \begin{tabular}{c l c c c c c c}
        \toprule
        \begin{tabular}{c}
            Dataset/ \\
            Architecture \\ 
        \end{tabular} &
        Optimiser & 
        \begin{tabular}{c}
            Train/Validation \\
            Accuracy (\%) 
        \end{tabular} &
        \begin{tabular}{c}
            Validation \\
            NLL 
        \end{tabular} &
        Epochs &
        \begin{tabular}{c}
        Time/ \\
        epoch (s) 
        \end{tabular} &
        ECE &
        AUROC \\
        \midrule
        \multirow{4}{*}{
            \begin{tabular}{c}
                CIFAR-10/ \\
                LeNet-5 \\
                (no DA) \\
            \end{tabular}}   
        & Adam          & 71.98 / \textbf{67.67} & \textbf{0.937} & 210 & 6.96 & \textbf{0.021} & 0.794 \\
        & BBB           & 66.84 / 64.61 & 1.018 & 800 & 11.43$^\dagger$ & 0.045 & 0.784 \\
        & MC-dropout    & 68.41 / \textbf{67.65} & 0.99 & 210 & 6.95 & 0.087 & \textbf{0.797} \\
        & VOGN          & 70.79 / \textbf{67.32} & \textbf{0.938} & 210 & 18.33 & 0.046 & \textbf{0.8} \\
        \midrule
        \multirow{3}{*}{
            \begin{tabular}{c}
                CIFAR-10/\\
                AlexNet\\
                (no DA)\\
            \end{tabular}}
        & Adam          & 100.0 / 67.94 & 2.83 & 161 & 3.12 & 0.262 & 0.793 \\
        & MC-dropout    & 97.56 / \textbf{72.20} & 1.077 & 160 & 3.25 & 0.140 & \textbf{0.818} \\
        & VOGN          & 79.07 / 69.03 & \textbf{0.93} & 160 & 9.98 & \textbf{0.024} & 0.796 \\
        \midrule
        \multirow{3}{*}{
            \begin{tabular}{c}
                CIFAR-10/\\
                AlexNet\\ 
            \end{tabular}}
        & Adam          & 97.92 / 73.59 & 1.480 & 161 & 3.08 & 0.262 & 0.793 \\
        & MC-dropout    & 80.65 / \textbf{77.04} & \textbf{0.667} & 160 & 3.20 & 0.114 & 0.828 \\
        & VOGN          & 81.15 / 75.48 & 0.703 & 160 & 10.02 & \textbf{0.016} & \textbf{0.832} \\
        \midrule
        \multirow{3}{*}{
            \begin{tabular}{c}
                CIFAR-10/\\
                ResNet-18\\ 
            \end{tabular}}
        & Adam          & 97.74 / \textbf{86.00} & 0.55 & 160 & 11.97 & 0.082 & \textbf{0.877} \\
        & MC-dropout    & 88.23 / 82.85 & 0.51 & 161 & 12.51 & 0.166 & 0.768 \\
        & VOGN          & 91.62 / 84.27 & \textbf{0.477} & 161 & 53.14 & \textbf{0.040} & \textbf{0.876} \\
        \midrule
        \multirow{6}{*}{
            \begin{tabular}{c}
                ImageNet/\\
                ResNet-18\\ 
            \end{tabular}}   
        & SGD           & 82.63 / \textbf{67.79} & \textbf{1.38} & 90 & 44.13  & 0.067 & 0.856 \\
        & Adam          & 80.96 / 66.39 & 1.44 & 90 & 44.40 & 0.064 & 0.855 \\
        & MC-dropout    & 72.96 / 65.64 & 1.43 & 90 & 45.86  & \textbf{0.012} & 0.856 \\
        & OGN          & 85.33 / 65.76 & 1.60 & 90 & 63.13 & 0.128 & 0.854 \\
        & VOGN          & 73.87 / \textbf{67.38} & \textbf{1.37} & 90 & 76.04 & 0.029 & 0.854 \\
        & K-FAC         & 83.73 / 66.58 & 1.493 & 60 & 133.69 & 0.158 & 0.842 \\
        & Noisy K-FAC   & 72.28 / 66.44 & 1.44 & 60 & 179.27 & 0.080 & 0.852 \\
        \bottomrule
    \end{tabular}
    \caption{
    Performance comparisons on different dataset/architecture combinations.
    Out of the 15 metrics (NLL, ECE, and AUROC on 5 dataset/architecture combinations), VOGN performs the best or tied best on 10 ,and is second-best on the other 5.
    Here DA means `Data Augmentation', NLL refers to `Negative Log Likelihood' (lower is better), ECE refers to `Expected Calibration Error' (lower is better), AUROC refers to `Area Under ROC curve' (higher is better). BBB is the Bayes By Backprop method. For ImageNet, the reported accuracy and negative log likelihood are the median value from the final 5 epochs. All hyperparameter settings are in Appendix \ref{app:vogn hyperparams}. See Table \ref{table:full results with std devs} for standard deviations. $^\dagger$ BBB is not parallelised (other methods have 4 processes), with 1 MC sample used for the convolutional layers (VOGN uses 6 samples per process).
    }
    \label{table:full results}
\end{table}

\subsection{Quality of the Predictive Probabilities}
\label{sec:uncertainty}

In this section, we compare the quality of the predictive probabilities for various methods. 
For Bayesian methods, we compute these probabilities by averaging over the samples from the posterior approximations (see Appendix \ref{app:uncertainty metrics} for details). For non-Bayesian methods, these are obtained using the point estimate of the weights.
We compare the probabilities using the following metrics: validation negative log-likelihood (NLL), area under ROC (AUROC) and expected calibration curves (ECE) \citep{naeini2015obtaining, guo2017oncalibration}. For the first and third metric, a lower number is better, while for the second, a higher number is better. 
See Appendix \ref{app:uncertainty metrics} for an explanation of these metrics.
Results are summarised in Table \ref{table:full results}. 
VOGN's uncertainty performance is more consistent and marginally better than the other methods, as expected from a more principled Bayesian method. 
Out of the 15 metrics (NLL, ECE and AUROC on 5 dataset/architecture combinations), VOGN performs the best or tied best on 10, and is second-best on the other 5. In contrast, both MC-dropout's and Adam's performance varies significantly, sometimes performing poorly, sometimes performing decently. MC-dropout is best on 4, and Adam is best on 1 (on LeNet-5; as argued earlier, the small architecture may result in underfitting). 
We also show calibration curves \citep{degroot1983comparison} in Figures \ref{fig:resnet18_imagenet_new} and \ref{fig:cc_cifar10}. 
Adam is consistently over-confident, with its calibration curve below the diagonal. Conversely, MC-dropout is usually under-confident. On ImageNet, MC-dropout performs well on ECE (all methods are very similar on AUROC), but this required an excessively tuned dropout rate (see Appendix \ref{app:MC Dropout sensitivity}).

\begin{figure}[!t]
    \centering
    \includegraphics[width=\textwidth]
    {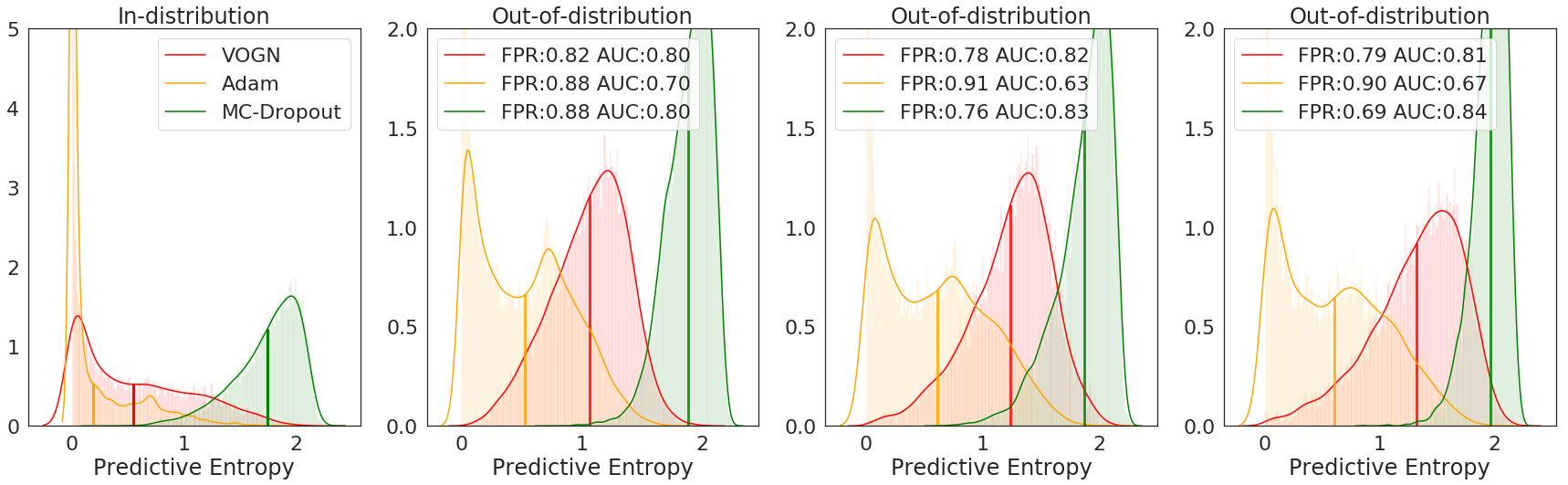}
    \caption{
    Histograms of predictive entropy for out-of-distribution tests for ResNet-18 trained on CIFAR-10.
    Going from left to right, the inputs are: the in-distribution dataset (CIFAR-10), followed by out-of-distribution data: SVHN, LSUN (crop), LSUN (resize). Also shown are the FPR at 95\% TPR metric (lower is better) and the AUROC metric (higher is better), averaged over 3 runs.
    We clearly see that VOGN's predictive entropy is generally low for in-distribution and high for out-of-distribution data, but this is not the case for other methods. Solid vertical lines indicate the mean predictive entropy.
    The standard deviations are small and therefore not reported.
    }
    \label{fig:resnet18 ood histograms}
\end{figure}

We also compare performance on out-of-distribution datasets.
When testing on datasets that are different from the training datasets, predictions should be more uncertain.
We use experimental protocol from the literature \citep{hendrycks2017baseline, lee2018training, devries2018learning, liang2018enhancing} to compare VOGN, Adam and MC-dropout on CIFAR-10.
We also borrow metrics from other works \citep{hendrycks2017baseline, lakshminarayanan2017simple}, showing predictive entropy histograms and also reporting AUROC and FPR at 95\% TPR.
See Appendix \ref{app:out-of-distribution results} for further details on the datasets and metrics.
Ideally, we want predictive entropy to be high on out-of-distribution data and low on in-distribution data.
Our results are summarised in Figure \ref{fig:resnet18 ood histograms} and Appendix \ref{app:out-of-distribution results}.
On ResNet-18 and AlexNet, VOGN's predictive entropy histograms show the desired behaviour: a spread of entropies for the in-distribution data, and high entropies for out-of-distribution data. 
Adam has many predictive entropies at zero, indicating Adam tends to classify out-of-distribution data too confidently. 
Conversely, MC-dropout's predictive entropies are generally high (particularly in-distribution), indicating MC-dropout has too much noise. 
On LeNet-5, we observe the same result as before: Adam and MC-dropout both perform well. The metrics (AUROC and FPR at 95\% TPR) do not provide a clear story across architectures.

\subsubsection{Performance on a Continual-learning task}
\label{subsubsec:continual_learning}

The goal of continual learning is to avoid forgetting of old tasks while sequentially observing new tasks. 
The past tasks are never visited again, making it difficult to remember them.
The field of continual learning has recently grown, with many approaches proposed to tackle this problem \citep{kirkpatrick2017overcoming, lopez2017gradient, nguyen2017variational, rusu2016progressive, schwarz2018progress}. Most approaches consider a simple setting where the tasks (such as classifying a subset of classes) arrive sequentially, and all the data from that task is available. We consider the same setup in our experiments. 

We compare to Elastic Weight Consolidation (EWC) \citep{kirkpatrick2017overcoming} and a VI-based approach called Variational Continual Learning (VCL) \citep{nguyen2017variational}. 
VCL employs BBB for each task, and we expect to boost its performance by replacing BBB by VOGN.
Figure \ref{fig:permuted mnist} shows results on a common benchmark called Permuted MNIST.
We use the same experimental setup as in \citet{swaroop2019improving}. 
In Permuted MNIST, each task consists of the entire MNIST dataset (10-way classification) with a different fixed random permutation applied to the input images' pixels. 
We run each method 20 times, with different random seeds for both the benchmark's permutations and model training. See Appendix \ref{app:continual learning details} for hyperparameter settings and further details. 
We see that VOGN performs at least as well as VCL, and far better than a popular approach called EWC \citep{kirkpatrick2017overcoming}. Additionally, as found in the batch learning setting, VOGN is much quicker than BBB: we run VOGN for only 100 epochs per task, whereas VCL requires 800 epochs per task to achieve best results \citep{swaroop2019improving}.

\section{Conclusions}
\label{sec:conclusions}

We successfully train deep networks with a natural-gradient variational inference method, VOGN, on a variety of architectures and datasets, even scaling up to ImageNet. 
This is made possible due to the similarity of VOGN to Adam, enabling us to boost performance by borrowing deep-learning techniques.
Our accuracies and convergence rates are comparable to SGD and Adam. Unlike them, however, VOGN retains the benefits of Bayesian principles, with well-calibrated uncertainty and good performance on out-of-distribution data. 
Better uncertainty estimates open up a whole range of potential future experiments, for example, small data experiments, active learning, adversarial experiments, and sequential decision making. Our results on a continual-learning task confirm this. Another potential avenue for research is to consider structured covariance approximations.

\newpage
{\bf Acknowledgements}

We would like to thank Hikaru Nakata (Tokyo Institute of Technology) and Ikuro Sato (Denso IT Laboratory, Inc.) for their help on the PyTorch implementation.
We are also thankful for the RAIDEN computing system and its support team at the RIKEN Center for AI Project which we used extensively for our experiments.
This research used computational resources of the HPCI system provided by Tokyo Institute of Technology (TSUBAME3.0) through the HPCI System Research Project (Project ID:hp190122).
K. O. is a Research Fellow of JSPS and is supported by JSPS KAKENHI Grant Number JP19J13477.

\bibliographystyle{plainnat}

\newpage
\appendix

\section{Noisy K-FAC algorithm}
\label{app:noisykfac}

Noisy K-FAC \citep{zhang2018noisy} attempts to approximate the structure of the full covariance matrix, and therefore the updates are a bit more involved than VOGN (see Equation \ref{eq:vogn}).
Assuming a fully-connected layer, we denote the weight matrix of layer by $\vW$. The Noisy K-FAC method estimates the parameters of a matrix-variate Gaussian distribution $q_t(\vW)=\mathcal{MN}(\vW|\vM_t,\vSigma_{2,t}\otimes\vSigma_{1,t})$ by using the following updates:
\begin{align}
    \vM_{t+1} &\leftarrow \vM_{t} - \alpha \sqr{\vA_{t+1}^\gamma}^{-1} \rnd{ \nabla_{W} E\left[\loss(y_i,f_W(\vx_i))\right] + \tilde{\delta} \vW_{t}} \sqr{\vS_{t+1}^\gamma}^{-1}, \\
    \vA_{t+1} &\leftarrow (1-\tilde{\beta}_t) \vA_{t} + \tilde{\beta}_t E\left[\va_{t}\va_{t}^\top\right], \quad
    \vS_{t+1} \leftarrow (1-\tilde{\beta}_t) \vS_{t} + \tilde{\beta}_t E\left[\vg_{t}\vg_{t}^\top\right] ,
\end{align}
where $\vW_{t}\sim q_t(\vW)$, $\vg_t := \nabla_{s} \loss(y_i, f_W(\vx_i))$ with $s=\vs_t := \vW_t^\top \va_t$, $\va_t$ is the input vector (the activation of the previous layer), $E\left[\cdot\right]$ is the average over the minibatch. $\tilde{\beta} := \beta\tau/N$, and $\gamma := \tilde{\gamma} + \gamma_{ex}$ with some \emph{external} damping factor $\gamma_{ex}$.
The covariance parameters are set to $\vSigma_{2,t}^{-1} := \tau \vA^{\gamma}_t/N$ and $\vSigma_{1,t}^{-1} := \vS_t^{\gamma}$, where $\vA^{\gamma}_t:=\vA_t+\pi_t\sqrt{\gamma}\vI$ and $\vS_t^{\gamma}:=\vS_t+\frac{1}{\pi_t}\sqrt{\gamma}\vI$. 
$\pi_t^2 (\pi_t>0)$ is the average eigenvalue of $\vA_t$ divided by that of $\vS_t$.
Similarly to the VOGN update in Equation \ref{eq:vogn}, the gradients are scaled by matrices $\vA_t$ and $\vS_t$, which are related to the precision matrix of the approximation. 

\section{Details on fast implementation of the Gauss-Newton approximation}
\label{app:GN implementation}
Current codebases are only optimised to directly return 
the average of gradients over the minibatch. 
In order to efficiently compute the Gauss-Newton (GN) approximation,
we modify the backward-pass to efficiently calculate 
the gradient per example in the minibatch, 
and extend the solution in \citet{goodfellow2015efficient} 
to both convolutional and batch normalisation layers. 

\subsection{Convolutional layer}
Consider a convolutional layer with a weight matrix $\vW\in\mathbb{R}^{C_{out}\times C_{in}k^2}$ 
(ignore bias for simplicity)
and an input tensor $\vA\in\mathbb{R}^{C_{in}\times H_{in} \times W_{in}}$, 
where $C_{out},C_{in}$ are the number of output, input channels, respectively, $H_{in},W_{in}$ are the spatial dimensions,
and $k$ is the kernel size. 
For any stride and padding, by applying \texttt{torch.nn.functional.unfold} function in PyTorch\footnote{\url{https://pytorch.org/docs/stable/nn.functional.html\#torch.nn.functional.unfold}},
we get the extended matrix $\vM_{A}\in\mathbb{R}^{C_{in}k^2\times H_{out}W_{out}}$ so that 
the output tensor $\vS$ is calculated by a matrix multiplication:
\begin{align}
    \vM_A
    &\leftarrow
    {\rm unfold}
    \left(
    \vA
    \right)
    \in\mathbb{R}^{C_{in}k^2\times H_{out}W_{out}}\,,
    \label{eq:unfold}
    \\
    \vM_S
    &\leftarrow
    \vW\vM_A
    \in\mathbb{R}^{C_{out}\times H_{out}W_{out}}\,,
    \\
    \vS
    &\leftarrow
    {\rm reshape}
    \left(
    \vM_S
    \right)
    \in\mathbb{R}^{C_{out}\times H_{out}\times W_{out}}\,,
\end{align}
where $H_{out},W_{out}$ are the spatial dimensions of the output feature map.
Using the matrix $\vM_A$, we can also get the gradient per example by a matrix multiplication:
\begin{align}
    \nabla_{M_S}\loss(y_i,f_W(\vx_i))
    &\leftarrow
    {\rm reshape}
    \left(
    \nabla_S \loss(y_i,f_W(\vx_i))
    \right)
    \in\mathbb{R}^{C_{out}\times H_{out}W_{out}}\,,
    \label{eq:reshape}
    \\
    \nabla_W \loss(y_i,f_W(\vx_i))
    &\leftarrow
    \nabla_{M_S}\loss(y_i,f_W(\vx_i))
    \vM_A^\top
    \in\mathbb{R}^{C_{out}\times C_{in}k^2}\,.
    \label{eq:mm_for_grad}
\end{align}
Note that in PyTorch, we can access to 
the inputs $\vA$ and 
the gradient 
$\nabla_S \loss(y_i,f_W(\vx_i))$
per example in the computational graph during a forward-pass and a backward-pass, respectively, by using the \texttt{Function Hooks}
\footnote{\url{https://pytorch.org/tutorials/beginner/former_torchies/nnft_tutorial.html\#forward-and-backward-function-hooks}}.
Hence, to get the gradient 
$\nabla_W \loss(y_i,f_W(\vx_i))$
per example, 
we only need to perform (\ref{eq:unfold}), (\ref{eq:reshape}), and (\ref{eq:mm_for_grad}) after the backward-pass for this layer.

\subsection{Batch normalisation layer}
Consider a batch normalisation layer follows a fully-connected layer, which activation is $\va\in\mathbb{R}^d$,
with the scale parameter $\vgamma\in\mathbb{R}^{d}$ 
and the shift parameter $\vbeta\in\mathbb{R}^d$,
we get the output of this layer $\vs\in\mathbb{R}^{d}$ by,
\begin{align}
    \vmu
    &\leftarrow
    E\left[
    \va
    \right]
    \in\mathbb{R}^d\,,
    \\
    \vsigma^2
    &\leftarrow
    E\left[
    \left(\va-\vmu\right)^2
    \right]
    \in\mathbb{R}^d\,,
    \\
    \hat{\va}
    &\leftarrow
    \frac{
        \va-\vmu
    }{
        \sqrt{\vsigma^2}
    }
    \in\mathbb{R}^d\,,
    \\
    \vs
    &\leftarrow
    \vgamma\hat{\va}+\vbeta
    \in\mathbb{R}^d\,,
\end{align}
where $E\left[\cdot\right]$ is the average over the minibatch and $\hat{\va}$ is the normalised input.
We can find the gradient with respect to parameters $\vgamma$ and $\vbeta$ per example by,
\begin{align}
    \nabla_{\gamma}\loss(y_i,f_W(\vx_i)
    &\leftarrow
    \nabla_{s}\loss(y_i,f_W(\vx_i)
    \circ\hat{\va}\,,
    \\
    \nabla_{\beta}\loss(y_i,f_W(\vx_i)
    &\leftarrow
    \nabla_{s}\loss(y_i,f_W(\vx_i)\,.
\end{align}
We can obtain the input $\va$ and the gradient $\nabla_{s}\loss(y_i,f_W(\vx_i)$ per example from the computational graph in PyTorch in the same way as a convolutional layer.

\subsection{Layer-wise block-diagonal Gauss-Newton approximation}
\begin{figure}
    \centering
    \includegraphics[width=0.9\textwidth]{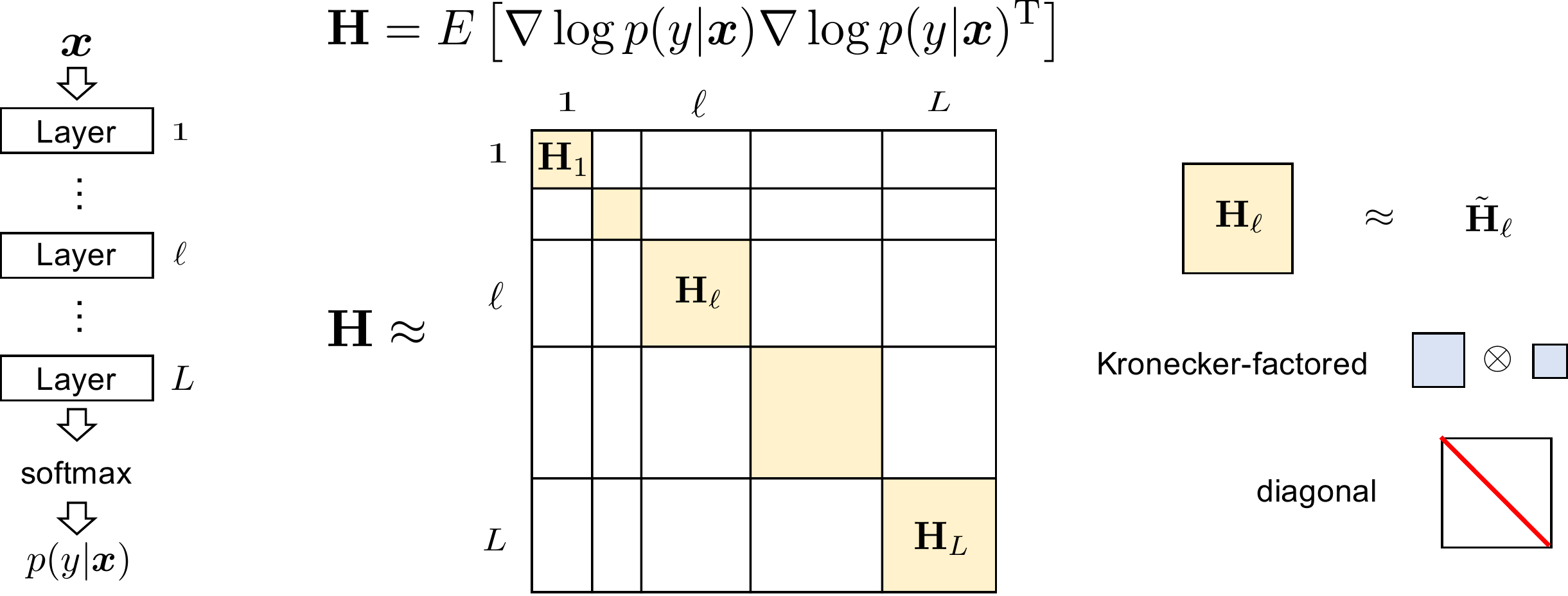}
    \caption{Layer-wise block-diagonal Gauss-Newton approximation}
    \label{fig:block_diagonal_gn}
\end{figure}

Despite using the method above, it is still intractable to compute the Gauss-Newton matrix (and its inverse) with respect to the weights of large-scale deep neural networks. We therefore apply two further approximations (Figure~\ref{fig:block_diagonal_gn}).
First, we view the Gauss-Newton matrix as a layer-wise block-diagonal matrix. This corresponds to ignoring the correlation between the weights of different layers. Hence for a network with $L$ layers, there are $L$ diagonal blocks, and $\matr{H}_{\ell}$ is the diagonal block corresponding to the $\ell$-th layer ($\ell=1,\dots,L$).
Second, we approximate each diagonal block $\matr{H}_{\ell}$ with $\tilde{\matr{H}}_{\ell}$, which is either a Kronecker-factored or diagonal matrix. Using a Kronecker-factored matrix as $\tilde{\matr{H}}_{\ell}$ corresponds to K-FAC; a diagonal matrix corresponds to a mean-field approximation in that layer.
By applying these two approximations, the update rule of the Gauss-Newton method can be written in a layer-wise fashion:
\begin{equation}
    \label{eq:layerwise_update}
    \boldsymbol{W}_{\ell,t+1}
    =
    \boldsymbol{W}_{\ell,t}
    -
    \alpha_t
    {\tilde{\matr{H}}_{\ell}(\boldsymbol{\theta}_t)}^{-1}
    \vec{g}_{\ell}(\boldsymbol{\theta}_t)
    \,\,\,
    (\ell=1,\dots,L)\,,
\end{equation}
where $\boldsymbol{W}_{\ell}$ is the weights in $\ell$-th layer, and
\begin{equation}
    \boldsymbol{\theta}
    =
    \left(
    \begin{array}{c c c c c}
        \rm{vec}(\boldsymbol{W}_1)^{\mathrm{T}} &
        \cdots &
        \rm{vec}(\boldsymbol{W}_{\ell})^{\mathrm{T}} &
        \cdots &
        \rm{vec}(\boldsymbol{W}_L)^{\mathrm{T}}
    \end{array}
    \right)^{\mathrm{T}}\,.
\end{equation}
Since the cost of computing $\tilde{\matr{H}}_{\ell}^{-1}$ 
is much cheaper compared to that of computing $\matr{H}^{-1}$,
our approximations make Gauss-Newton much more practical in deep learning.

In the distributed setting (see Figure \ref{fig:distributed_vi}), each parallel process (corresponding to 1 GPU) calculates the GN matrix for its local minibatch. Then, one GPU adds them together and calculates the inverse. This inversion step can also be parallelised after making the block-diagonal approximation to the GN matrix. After inverting the GN matrix, the standard deviation $\vsigma$ is updated (line 9 in Algorithm \ref{alg:vogn}), and sent to each parallel process, allowing each process to draw independently from the posterior.

In the Noisy K-FAC case, a similar distributed scheme is used, except each parallel process now has both matrices $\vS$ and $\vA$ (see Appendix \ref{app:noisykfac}). When using K-FAC approximations to the Gauss-Newton blocks for other layers, \citet{osawa2018secondorder} empirically showed that the BatchNorm layer can be approximated with a diagonal matrix without loss of accuracy, and we find the same. We therefore use diagonal $\tilde{\matr{H}}_{\ell}$ with K-FAC and Noisy K-FAC in BatchNorm layers (see Table \ref{table:layer_approximations}). For further details on how to efficiently parallelise K-FAC in the distributed setting, please see \citet{osawa2018secondorder}.

\begin{table}[htbp]
    \centering
    \begin{tabular}{c c c c}
        \toprule
        optimiser & convolution & fully-connected & Batch Normalisation \\
        \midrule
        OGN & diagonal & diagonal & diagonal \\ 
        VOGN & diagonal & diagonal & diagonal \\ 
        K-FAC & Kronecker-factored & Kronecker-factored & diagonal \\ 
        Noisy K-FAC & Kronecker-factored & Kronecker-factored & diagonal \\ 
        \bottomrule
    \end{tabular}
    \caption{The approximation used for each layer type's diagonal block $\tilde{\matr{H}}_{\ell}$ for the different optimisers tested this paper.}
    \label{table:layer_approximations}
\end{table}

\begin{sidewaystable}
    \centering
    \footnotesize
    \begin{tabular}{c l c c c c c c c c}
        \toprule
        \begin{tabular}{c}
            Dataset/ \\
            Architecture \\ 
        \end{tabular} &
        Optimiser &
        \begin{tabular}{c}
            Train \\
            Acc (\%) 
        \end{tabular} &
        \begin{tabular}{c}
            Train \\
            NLL
        \end{tabular} &
        \begin{tabular}{c}
            Validation \\
            Acc (\%) 
        \end{tabular} &
        \begin{tabular}{c}
            Validation \\
            NLL 
        \end{tabular} & 
        ECE &
        AUROC &
        Epochs &
        \begin{tabular}{c}
        Time/ \\
        epoch (s) 
        \end{tabular} \\
        \midrule
        \multirow{4}{*}{
            \begin{tabular}{c}
                CIFAR-10/ \\
                LeNet-5 \\
                (no DA) \\
            \end{tabular}}   
        & Adam          & 71.98 $\pm$ 0.117 & 0.733 $\pm$ 0.021 & \textbf{67.67} $\pm$ \textbf{0.513} & \textbf{0.937} $\pm$ \textbf{0.012} & \textbf{0.021} $\pm$ \textbf{0.002} & 0.794 $\pm$ 0.001 & 210 & 6.96 \\
        & BBB           & 66.84 $\pm$ 0.003 & 0.957 $\pm$ 0.006 & 64.61 $\pm$ 0.331 & 1.018  $\pm$ 0.006 & 0.045 $\pm$ 0.005 & 0.784 $\pm$ 0.003 & 800 & 11.43 \\
        & MC-dropout    & 68.41 $\pm$ 0.581 & 0.870 $\pm$ 0.101 & \textbf{67.65} $\pm$ \textbf{1.317} & 0.99 $\pm$ 0.026 & 0.087 $\pm$ 0.009 & \textbf{0.797} $\pm$ \textbf{0.006} & 210 & 6.95 \\
        & VOGN          & 70.79 $\pm$ 0.763 & 0.880 $\pm$ 0.02 & \textbf{67.32} $\pm$ \textbf{1.310} & \textbf{0.938} $\pm$ \textbf{0.024} & 0.046 $\pm$ 0.002 & \textbf{0.8} $\pm$ \textbf{0.002} & 210 & 18.33 \\
        \midrule
        \multirow{3}{*}{
            \begin{tabular}{c}
                CIFAR-10/\\
                AlexNet\\
                (no DA)\\
            \end{tabular}}
        & Adam          & 100.0 $\pm$ 0 & 0.001 $\pm$ 0 & 67.94 $\pm$ 0.537 & 2.83 $\pm$ 0.02 & 0.262 $\pm$ 0.005 & 0.793 $\pm$ 0.001 & 161 & 3.12 \\
        & MC-dropout    & 97.56 $\pm$ 0.278 & 0.058 $\pm$ 0.014 & \textbf{72.20} $\pm$ \textbf{0.177} & 1.077 $\pm$ 0.012 & 0.140 $\pm$ 0.004 & \textbf{0.818} $\pm$ \textbf{0.002} & 160 & 3.25 \\
        & VOGN          & 79.07 $\pm$ 0.248 & 0.696 $\pm$ 0.020 & 69.03 $\pm$ 0.419 & \textbf{0.931} $\pm$ \textbf{0.017} & \textbf{0.024} $\pm$ \textbf{0.010} & 0.796 $\pm$ 0 & 160 & 9.98 \\
        \midrule
        \multirow{3}{*}{
            \begin{tabular}{c}
                CIFAR-10/\\
                AlexNet\\ 
            \end{tabular}}
        & Adam          & 97.92 $\pm$ 0.140 & 0.057 $\pm$ 0.006 & 73.59 $\pm$ 0.296 & 1.480 $\pm$ 0.015 & 0.262 $\pm$ 0.005 & 0.793 $\pm$ 0.001& 161 & 3.08 \\
        & MC-dropout    & 80.65 $\pm$ 0.615 & 0.47 $\pm$ 0.052 & \textbf{77.04} $\pm$ \textbf{0.343} & \textbf{0.667} $\pm$ \textbf{0.012} & 0.114 $\pm$ 0.002 & 0.828 $\pm$ 0.002 & 160 & 3.20 \\
        & VOGN          & 81.15 $\pm$ 0.259 & 0.511 $\pm$ 0.039 & 75.48 $\pm$ 0.478 & 0.703 $\pm$ 0.006 & \textbf{0.016} $\pm$ \textbf{0.001} & \textbf{0.832} $\pm$ \textbf{0.002} & 160 & 10.02 \\
        \midrule
        \multirow{3}{*}{
            \begin{tabular}{c}
                CIFAR-10/\\
                ResNet-18\\ 
            \end{tabular}}
        & Adam          & 97.74 $\pm$ 0.140 & 0.059 $\pm$ 0.012 & \textbf{86.00} $\pm$ \textbf{0.257} & 0.55 $\pm$ 0.01 & 0.082 $\pm$ 0.002 & \textbf{0.877} $\pm$ \textbf{0.001} & 160 & 11.97 \\
        & MC-dropout    & 88.23 $\pm$ 0.243 & 0.317 $\pm$ 0.045 & 82.85 $\pm$ 0.208 & 0.51 $\pm$ 0 & 0.166 $\pm$ 0.025 & 0.768 $\pm$ 0.004 & 161 & 12.51 \\
        & VOGN          & 91.62 $\pm$ 0.07 & 0.263 $\pm$ 0.051 & 84.27 $\pm$ 0.195 & \textbf{0.477} $\pm$ \textbf{0.006} & \textbf{0.040} $\pm$ \textbf{0.002} & \textbf{0.876} $\pm$ \textbf{0.002} & 161 & 53.14 \\
        \midrule
        \multirow{7}{*}{
            \begin{tabular}{c}
                ImageNet/\\
                ResNet-18\\ 
            \end{tabular}}   
        & SGD           & 82.63 $\pm$ 0.058 & 0.675 $\pm$ 0.017 & \textbf{67.79} $\pm$ \textbf{0.017} & \textbf{1.38} $\pm$ \textbf{0} & 0.067 & 0.856 & 90 & 44.13 \\
        & Adam          & 80.96 $\pm$ 0.098 & 0.723 $\pm$ 0.015 & 66.39 $\pm$ 0.168 & 1.44 $\pm$ 0.01 & 0.064 & 0.855 & 90 & 44.40 \\
        & MC-dropout    & 72.96 & 1.12 & 65.64 & 1.43 & \textbf{0.012} & 0.856 & 90 & 45.86 \\
        & OGN          & 85.33 $\pm$ 0.057 & 0.526 $\pm$ 0.005 & 65.76 $\pm$ 0.115 & 1.60 $\pm$ 0.00 & 0.128 $\pm$ 0.004 & 0.8543 $\pm$ 0.001 & 90 & 63.13 \\
        & VOGN          & 73.87 $\pm$ 0.061 & 1.02 $\pm$ 0.01 & \textbf{67.38} $\pm$ \textbf{0.263} & \textbf{1.37} $\pm$ \textbf{0.01} & 0.0293 $\pm$ 0.001 & 0.8543 $\pm$ 0 & 90 & 76.04 \\
        & K-FAC         & 83.73 $\pm$ 0.058 & 0.571 $\pm$ 0.016 & 66.58 $\pm$ 0.176    & 1.493 $\pm$ 0.006 & 0.158 $\pm$ 0.005 & 0.842 $\pm$ 0.005 & 60 & 133.69 \\
        & Noisy K-FAC   & 72.28 & 1.075 & 66.44 & 1.44 & 0.080 & 0.852 & 60 & 179.27 \\
        \bottomrule
    \end{tabular}
    \caption{
    Comparing optimisers on different dataset/architecture combinations, means and standard deviations over three runs. DA: Data Augmentation, Acc.: Accuracy (higher is better), NLL: Negative Log Likelihood (lower is better), ECE: Expected Calibration Error (lower is better), AUROC: Area Under ROC curve (higher is better), BBB: Bayes By Backprop. For ImageNet results, the reported accuracy and negative log likelihood are the median value from the final 5 epochs.
    }
    \label{table:full results with std devs}
\end{sidewaystable}

\section{OGN: A deterministic version of VOGN}
\label{app:ogn}
To easily apply the tricks and techniques of deep-learning methods, we recommend to first test them on a deterministic version of VOGN, which we call the online Gauss-Newton (OGN) method.
In this method, we approximate the gradients at the mean of the Gaussian, rather than using MC samples\footnote{This gradient approximation used here is referred to as the \emph{zeroth-order delta approximation} where  $\myexpect_{q}[\hat{\vg}(\vparam)] \approx \hat{\vg}(\vmu)$ (see Appendix A.6 in \citet{khan2012variational} for details).}.
This results in an update without any sampling, as shown below (we have replaced $\vmu_t$ by $\vparam_t$ since there is no distinction between them): 
\begin{align}
    \vw_{t+1} &\leftarrow \vparam_t - \alpha_t \frac{\hat{\vg}(\vparam_t) + \tilde{\delta}\vw_t}{\vs_{t+1} + \tilde{\delta}} , \quad
    \vs_{t+1} \leftarrow (1- \tau\beta_t) \vs_{t}+ \beta_t \frac{1}{M} \sum_{i\in\minibatch_t} \rnd{\vg_i(\vparam_t)}^2.
    \label{eq:ogn}
\end{align}
At each iteration, we still get a Gaussian $\gauss(\vparam|\vparam_t, \hat{\vSigma}_t)$ with $\hat{\vSigma}_t := \diag(1/(N(\vs_t + \tilde{\delta} )))$.
It is easy to see that, like SG methods, this algorithm converges to a local minima of the loss function, thereby obtaining a Laplace approximation instead of a variational approximation.
The advantage of OGN is that this can be used as a stepping stone, when switching from Adam to VOGN. Since it does not involve sampling, the tricks and techniques applied to Adam are easier to apply to OGN than VOGN. However, due to the lack of averaging over samples, this algorithm is less effective to preserve the benefits of Bayesian principles, and gives slightly worse uncertainty estimates.

\section{Hyperparameter settings}
\label{app:vogn hyperparams}
\begin{figure}[ht!]
    \centering
    \begin{minipage}{\textwidth}
        \centering
        \includegraphics[scale=0.25]{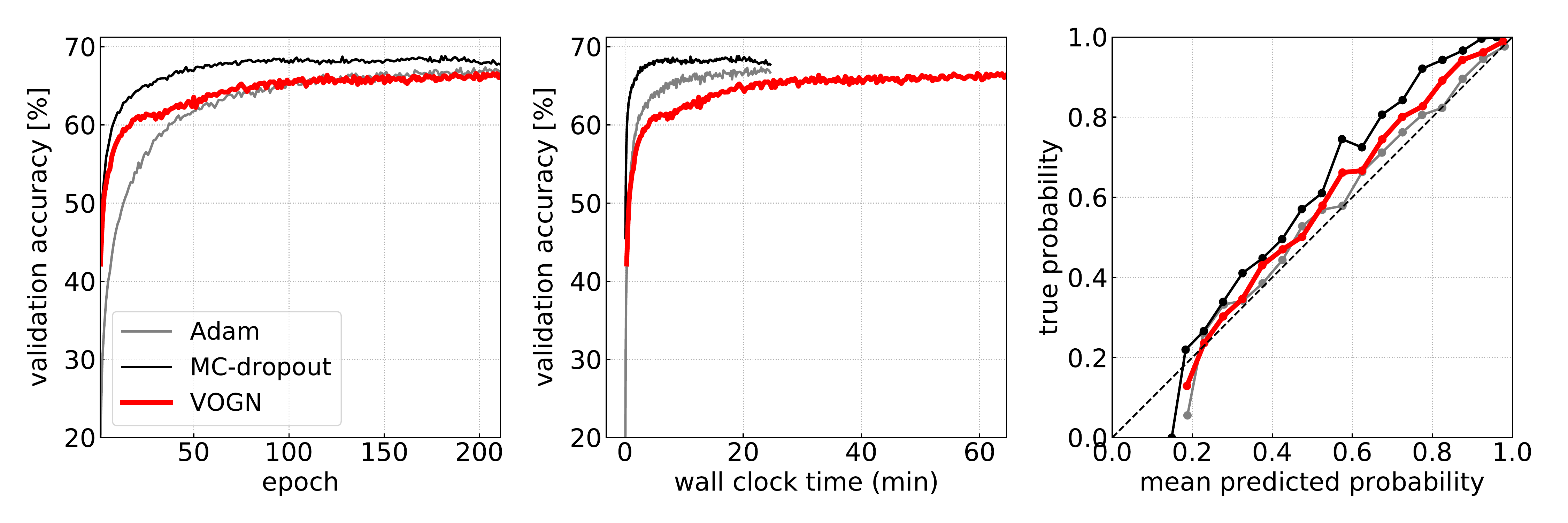}
        \subcaption{LeNet-5 on CIFAR-10 (no DA)}
        \label{fig:all_curves_lenet_noda_cifar10}
    \end{minipage}
    \begin{minipage}{\textwidth}
        \centering
        \includegraphics[scale=0.25]{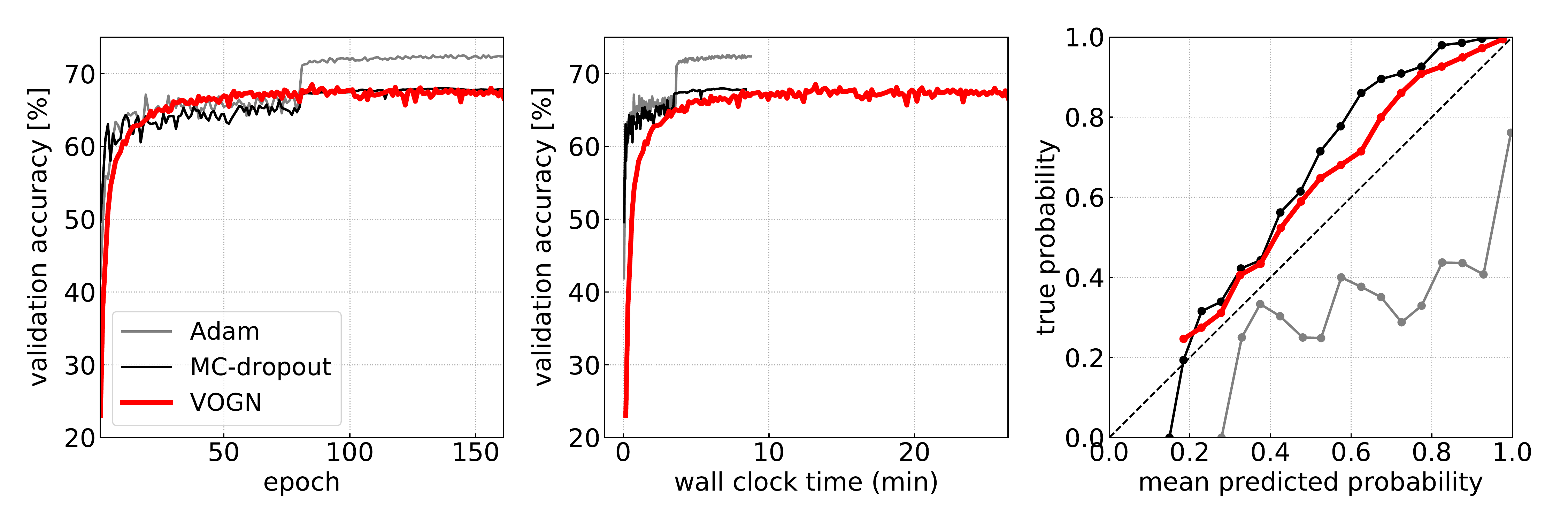}
        \subcaption{AlexNet on CIFAR-10 (no DA)}
        \label{fig:all_curves_alexnet_noda_cifar10}
    \end{minipage}
    \begin{minipage}{\textwidth}
        \centering
        \includegraphics[scale=0.25]{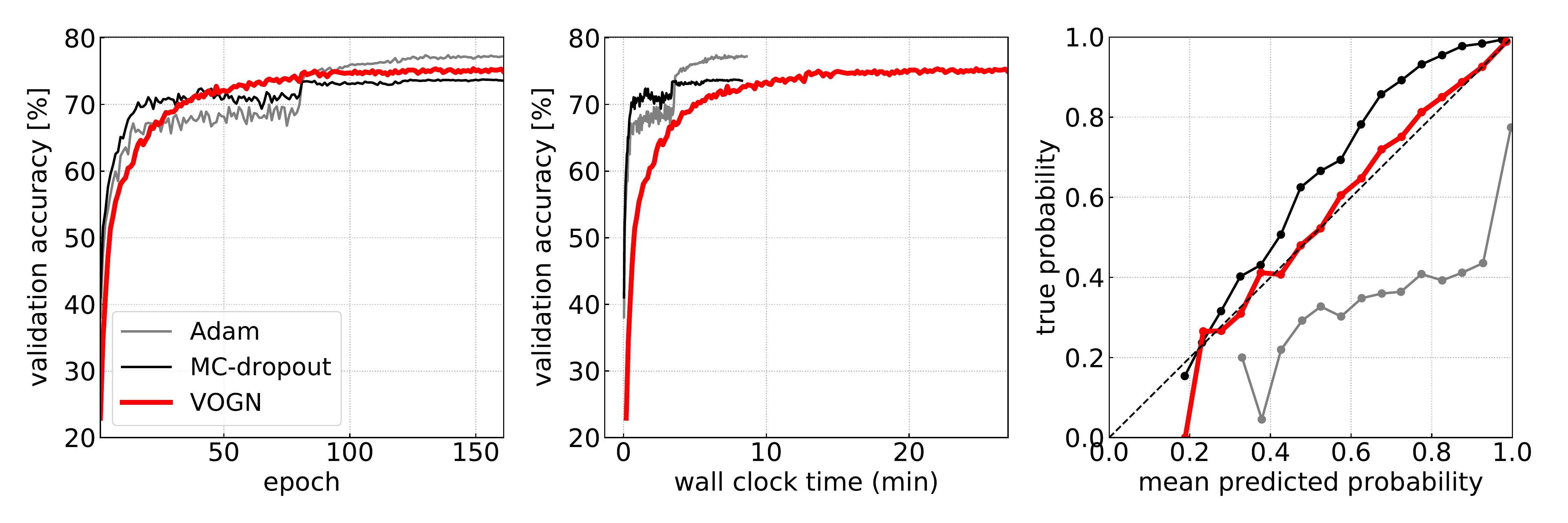}
        \subcaption{AlexNet on CIFAR-10}
        \label{fig:all_curves_alexnet_cifar10}
    \end{minipage}
    \begin{minipage}{\textwidth}
        \centering
        \includegraphics[scale=0.25]{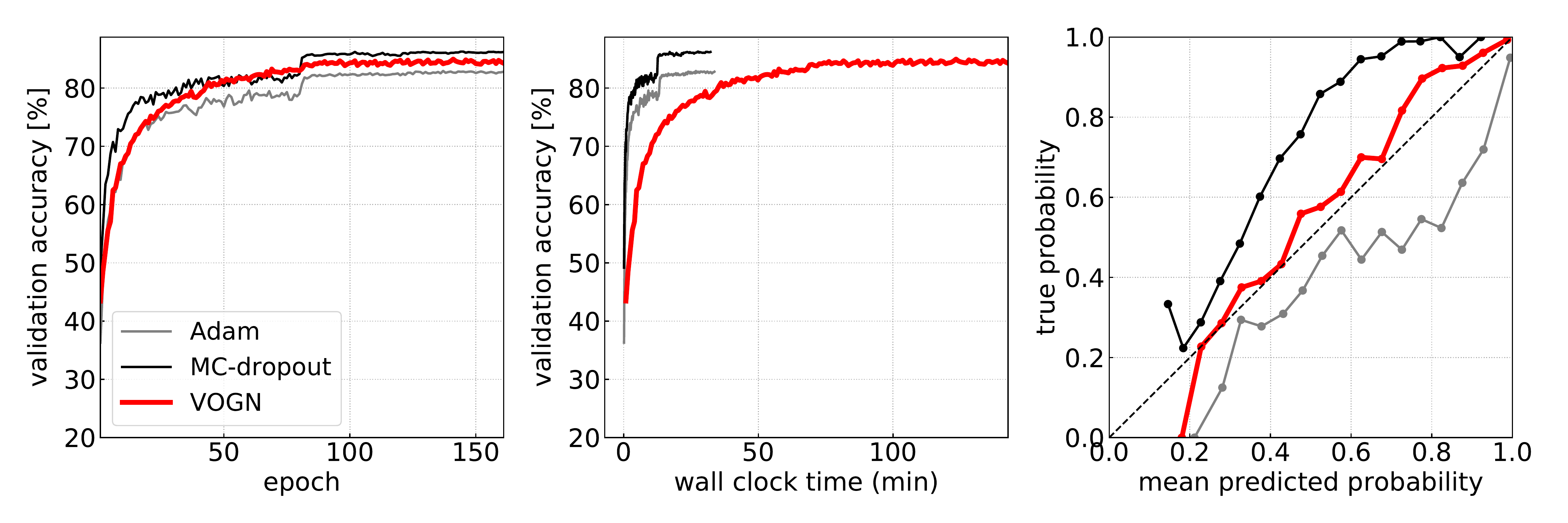}
        \subcaption{ResNet-18 on CIFAR-10}
        \label{fig:all_curves_resnet18_cifar10}
    \end{minipage}
    \begin{minipage}{\textwidth}
        \centering
        \includegraphics[scale=0.25]{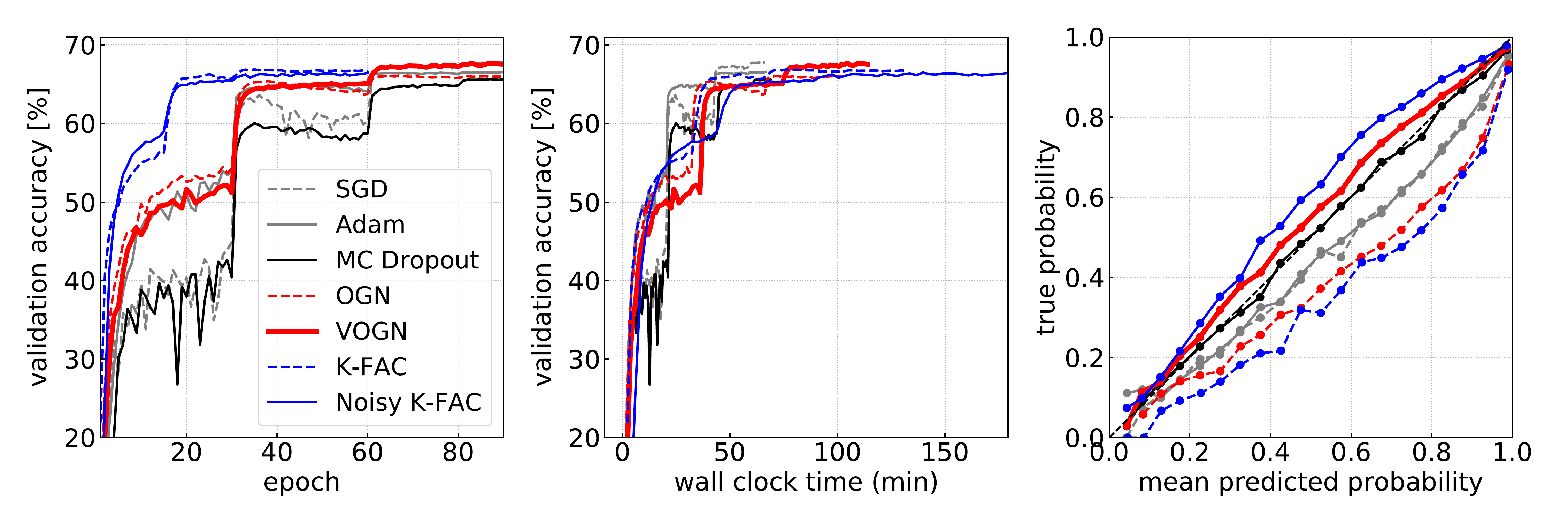}
        \subcaption{ResNet-18 on ImageNet}
        \label{fig:all_curves_resnet18_imagenet}
    \end{minipage}
    \caption{All results in Table~\ref{table:full results}}
    \label{fig:all_curves}
\end{figure}

Hyperparameters for all results shown in Table \ref{table:full results} are given in Table \ref{table:hyperparams}. The settings for distributed VI training are given in Table \ref{table:settings}.
Please see \citet{goyal2017accurate} and \citet{osawa2018secondorder} for best practice on these hyperparameter values.

\begin{table}[thbp]
    \centering
    \begin{adjustbox}{max width=\textwidth}
    \footnotesize
    \begin{tabular}{c l c c c c c c c}
        \toprule
        \begin{tabular}{c}
            Dataset/ \\
            Architecture \\ 
        \end{tabular} &
        Optimiser &
        \begin{tabular}{c}
            $\alpha_{init}$
        \end{tabular} &
        \begin{tabular}{c}
            $\alpha$
        \end{tabular} &
        \begin{tabular}{c}
            Epochs to decay $\alpha$ 
        \end{tabular} &
        \begin{tabular}{c}
            $\beta_1$
        \end{tabular} & 
            $\beta_2$ &
        \begin{tabular}{c}
            Weight \\
            decay
        \end{tabular} &
        \begin{tabular}{c}
            L2 reg
        \end{tabular} \\
        \midrule
        \multirow{4}{*}{
            \begin{tabular}{c}
                CIFAR-10/ \\
                LeNet-5 \\
                (no DA) \\
            \end{tabular}}   
        & Adam          & - & 1e-3 & - & 0.1 & 0.001 & 1e-2 & - \\
        & BBB           & - & 1e-3 & - & 0.1 & 0.001 & - & - \\
        & MC-dropout    & - & 1e-3 & - & 0.9 & - & - & 1e-4 \\
        & VOGN          & - & 1e-2 & - & 0.9 & 0.999 & - & - \\
        \midrule
        \multirow{3}{*}{
            \begin{tabular}{c}
                CIFAR-10/\\
                AlexNet\\
                (no DA)\\
            \end{tabular}}
        & Adam          & - & 1e-3 & [80, 120] & 0.1 & 0.001 & 1e-4 & - \\
        & MC-dropout    & - & 1e-1 & [80, 120] & 0.9 & - & - & 1e-4 \\
        & VOGN          & - & 1e-4 & [80, 120] & 0.9 & 0.999 & - & - \\
        \midrule
        \multirow{3}{*}{
            \begin{tabular}{c}
                CIFAR-10/\\
                AlexNet\\ 
            \end{tabular}}
        & Adam          & - & 1e-3 & [80, 120] & 0.1 & 0.001 & 1e-4 & - \\
        & MC-dropout    & - & 1e-1 & [80, 120] & 0.9 & - & - & 1e-4 \\
        & VOGN          & - & 1e-4 & [80, 120] & 0.9 & 0.999 & - & - \\
        \midrule
        \multirow{3}{*}{
            \begin{tabular}{c}
                CIFAR-10/\\
                ResNet-18\\ 
            \end{tabular}}
        & Adam          & - & 1e-3 & [80, 120] & 0.1 & 0.001 & 5e-4 & - \\
        & MC-dropout    & - & 1e-1 & [80, 120] & 0.9 & - & - & 1e-4 \\
        & VOGN          & - & 1e-4 & [80, 120] & 0.9 & 0.999 & - & - \\
        \midrule
        \multirow{7}{*}{
            \begin{tabular}{c}
                ImageNet/\\
                ResNet-18\\ 
            \end{tabular}}   
        & SGD           & 1.25e-2 & 1.6 & [30, 60, 80] & 0.9 & - & - & 1e-4 \\
        & Adam          & 1.25e-5 & 1.6e-3 & [30, 60, 80] & 0.1 & 0.001 & 1e-4 & - \\
        & MC-dropout    & 1.25e-2 & 1.6 & [30, 60, 80] & 0.9 & - & - & 1e-4 \\
        & OGN          & 1.25e-5 & 1.6e-3 & [30, 60, 80] & 0.9 & 0.9 & - & 1e-5 \\
        & VOGN          & 1.25e-5 & 1.6e-3 & [30, 60, 80] & 0.9 & 0.999 & - & - \\
        & K-FAC         & 1.25e-5 & 1.6e-3 & [15, 30, 45] & 0.9 & 0.9 & - & 1e-4 \\
        & Noisy K-FAC   & 1.25e-5 & 1.6e-3 & [15, 30, 45] & 0.9 & 0.9 & - & - \\
        \bottomrule
    \end{tabular}
    \end{adjustbox}
    \caption{
    Hyperparameters for all results in Table \ref{table:full results}
    }
    \label{table:hyperparams}
\end{table}

\begin{table}[thbp]
    \centering
    \begin{adjustbox}{max width=\textwidth}
    \footnotesize
    \begin{tabular}{c l c c c c c c c c c}
        \toprule
        Optimiser &
        \begin{tabular}{c}
            Dataset/ \\
            Architecture \\ 
        \end{tabular} &
        \begin{tabular}{c}
            $M$
        \end{tabular} &
        \begin{tabular}{c}
           $\#$ GPUs
        \end{tabular} &
        \begin{tabular}{c}
            $K$
        \end{tabular} &
        \begin{tabular}{c}
            $\tau$
        \end{tabular} &
        \begin{tabular}{c}
            $\rho$
        \end{tabular} &
        \begin{tabular}{c}
            $N_{orig}$
        \end{tabular} &
        \begin{tabular}{c}
            $\delta$
        \end{tabular} &
        \begin{tabular}{c}
            $\tilde{\delta}$
        \end{tabular} &
        \begin{tabular}{c}
            $\gamma$
        \end{tabular} \\
        \midrule
        \multirow{5}{*}{
            VOGN 
        } 
        & \begin{tabular}{c}
                CIFAR-10/ \\
                LeNet-5 \\
                (no DA) \\
        \end{tabular}      
        & 128 & 4 & 6 & 0.1 $\rightarrow$ 1 & 1 & 50,000 & 100 & 2e-4 $\rightarrow$ 2e-3 & 1e-3 \\
        & \begin{tabular}{c}
                CIFAR-10/\\
                AlexNet\\
                (no DA)\\
        \end{tabular}      
        & 128 & 8 & 3 & 0.05 $\rightarrow$ 1 & 1 & 50,000 & 0.5 & 5e-7 $\rightarrow$ 1e-5 & 1e-3 \\
        & \begin{tabular}{c}
                CIFAR-10/\\
                AlexNet\\
        \end{tabular} 
        & 128 & 8 & 3 & 0.5 $\rightarrow$ 1 & 10 & 50,000 & 0.5 & 5e-7 $\rightarrow$ 1e-5 & 1e-3 \\
        & \begin{tabular}{c}
                CIFAR-10/\\
                ResNet-18\\
                \end{tabular}
        & 256 & 8 & 5 & 1 & 10 & 50,000 & 50 & 1e-3 & 1e-3 \\
        & \begin{tabular}{c}
                ImageNet/\\
                ResNet-18\\ 
        \end{tabular}        
        & 4096 & 128 & 1 & 1 & 5 & 1,281,167 & 133.3 & 2e-5 & 1e-4 \\
        \midrule
        \multirow{1}{*}{
            Noisy K-FAC
        }
        & \begin{tabular}{c}
                ImageNet/\\
                ResNet-18\\  
        \end{tabular}
        & 4096 & 128 & 1 & 1 & 5 & 1,281,167 & 133.3 & 2e-5 & 1e-4 \\
        \bottomrule
    \end{tabular}
    \end{adjustbox}
    \caption{
    Settings for distributed VI training
    }
    \label{table:settings}
\end{table}

\subsection{Bayes by Backprop for CIFAR-10/LeNet-5 training}

We use hyperparameter settings and training procedure for Bayes by Backprop (BBB) \citep{blundell2015weight} as suggested by \citet{swaroop2019improving}. This includes using the local reparameterisation trick, initialising means and variances at small values, using 10 MC samples per minibatch during training for linear layers (1 MC sample per minibatch for convolutional layers) and 100 MC samples per minibatch during testing for linear layers (10 MC samples per minibatch for convolutional layers). Note that BBB has twice as many parameters to optimise than Adam or SGD (means and variances for each weight in the deep neural network). The fewer MC samples per minibatch for convolutional layers speed up training time per epoch while empirically not reducing convergence rate.

\subsection{Continual learning experiment}
\label{app:continual learning details}

Following the setup of \citet{swaroop2019improving}, all models are run with two hidden layers, of 100 hidden units each, with ReLU activation functions. VCL is run with the same hyperparameter settings as in \citet{swaroop2019improving}. We perform a grid search over EWC's $\lambda$ hyperparameter, finding that $\lambda=100$ performs the best, exactly like in \citet{nguyen2017variational}. 

VOGN is run for 100 epochs per task. Parameters are initialised before training with the default PyTorch initialisation for linear layers. The initial precision is 1e6. A standard normal initial prior is used, just like in VCL. Between tasks, the mean and precision are initialised in the same way as for the first task. The learning rate $\alpha = 1e-3$, the batch size $M = 256$, $\beta_1 = 0$, $\beta_2 = 1e-3$, 10 MC samples are used during training and 100 for testing.
We run each method 20 times, with different random seeds for both the benchmark's permutation and for model training.

\section{Effect of prior variance and dataset size reweighting factor}
\label{app:prior variance}

\begin{figure}[h]
    \centering
    \includegraphics[width=\textwidth]{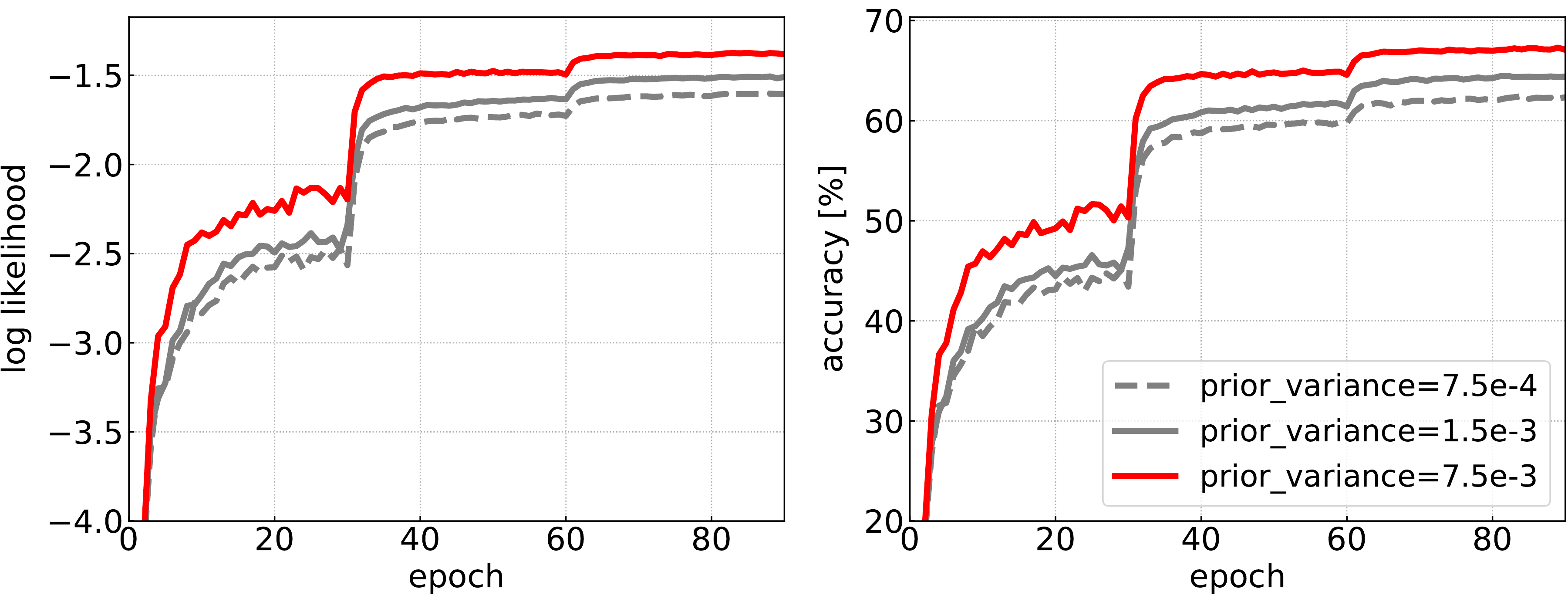}
    \caption{Effect of prior variance on VOGN training ResNet-18 on ImageNet.}
    \label{fig:resnet18_imagenet_vogn_prior_var}
\end{figure}

\begin{figure}[h]
    \centering
    \includegraphics[width=\textwidth]{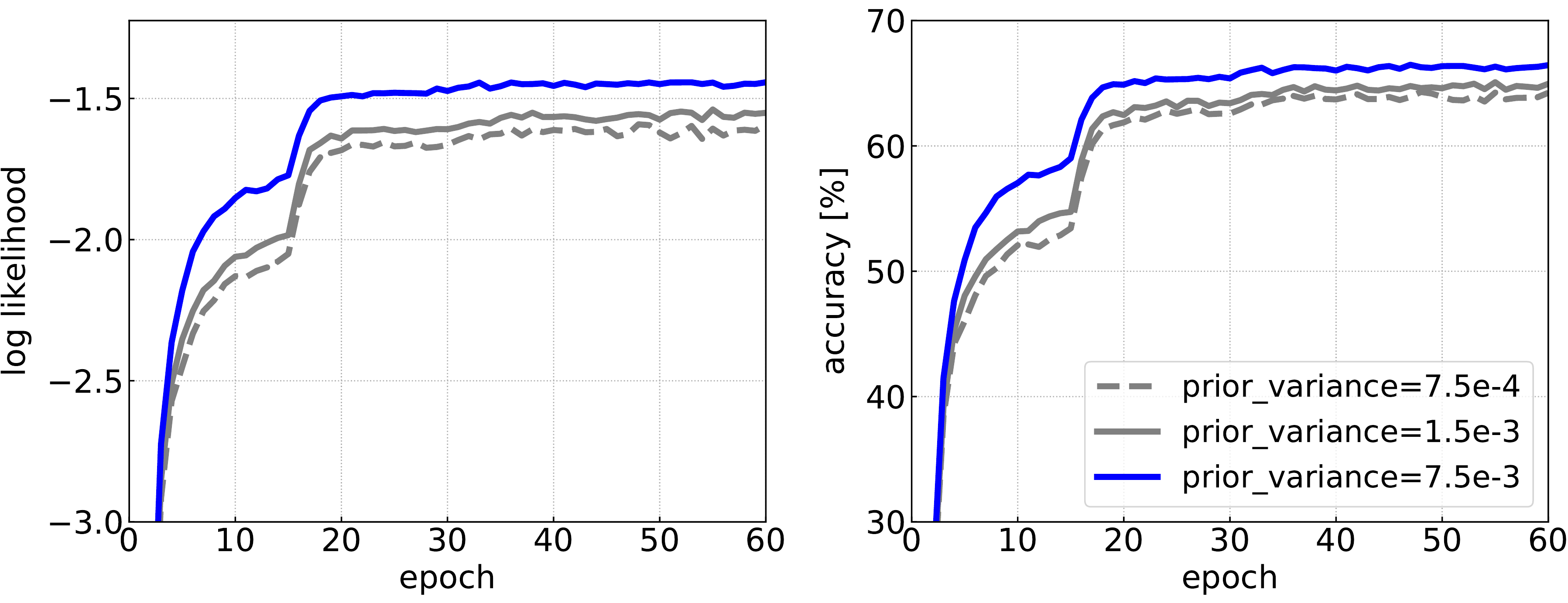}
    \caption{Effect of prior variance on Noisy K-FAC training ResNet-18 on ImageNet.}
    \label{fig:resnet18_imagenet_noisy_kfac_prior_var}
\end{figure}

We show the effect of changing the prior variance ($\delta^{-1}$ in Algorithm \ref{alg:vogn}) in Figures \ref{fig:resnet18_imagenet_vogn_prior_var} and \ref{fig:resnet18_imagenet_noisy_kfac_prior_var}. We can see that increasing the prior variance improves validation performance (accuracy and log likelihood). However, increasing prior variance also always increases the train-test gap, without exceptions, when the other hyperparameters are held constant. As an example, training VOGN on ResNet-18 on ImageNet with a prior variance of $7.5e-4$ has train-test accuracy and log likelihood gaps of 2.29 and 0.12 respectively. When the prior variance is increased to $7.5e-3$, the respective train-test gaps increase to 6.38 and 0.34 (validation accuracy and validation log likelihood also increase, see Figure \ref{fig:resnet18_imagenet_vogn_prior_var}).

With increased prior variance, VOGN (and Noisy K-FAC) reach converged solutions more like their non-Bayesian counterparts, where overfitting is an issue. This is as expected from Bayesian principles.

\begin{figure}[h]
    \centering
    \includegraphics[width=\textwidth]{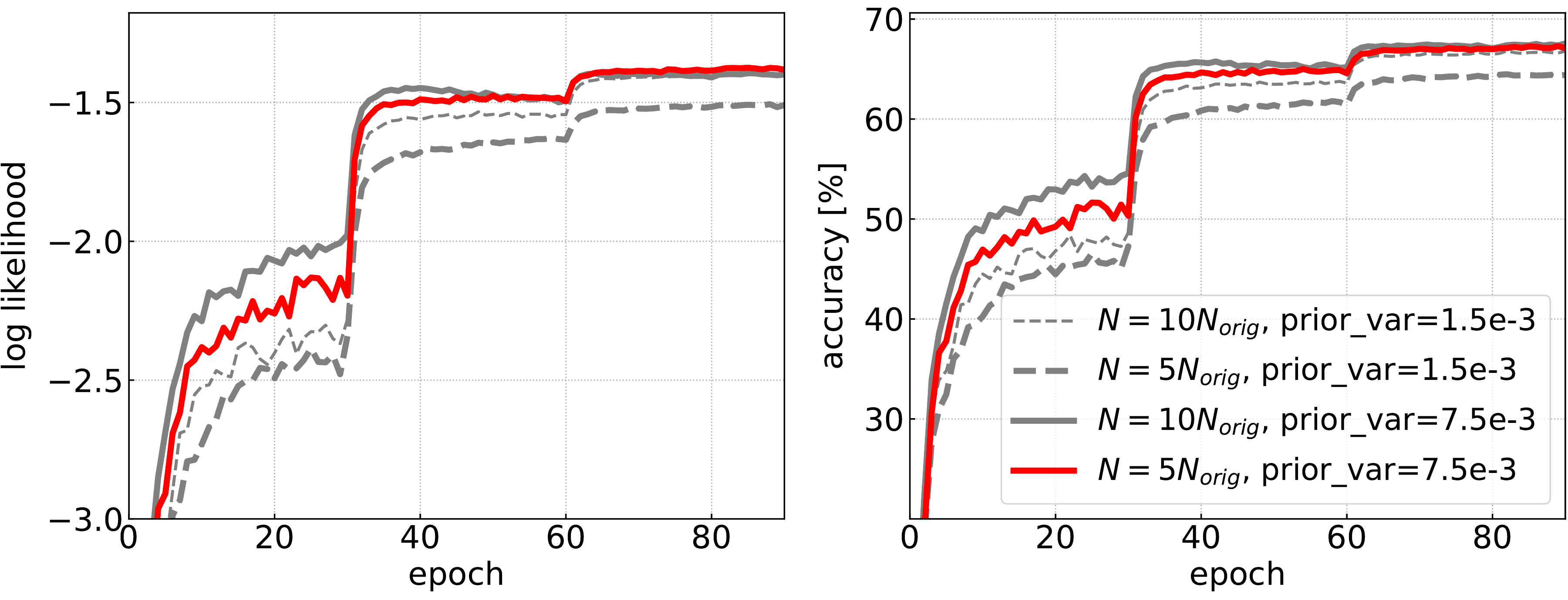}
    \caption{Effect of changing the dataset size reweighting factor $\rho$ and prior variance on VOGN training ResNet-18 on ImageNet.}
    \label{fig:resnet18_imagenet_vogn_datasetsize_priorvar}
\end{figure}

Figure \ref{fig:resnet18_imagenet_vogn_datasetsize_priorvar} shows the combined effect of the dataset reweighting factor $\rho$ and prior variance. When $\rho$ is set to a value in the correct order of magnitude, it does not affect performance so much: instead, we should tune $\delta$. This is our methodology when dealing with $\rho$. Note that we set $\rho$ for ImageNet to be smaller than that for CIFAR-10 because the data augmentation cropping step uses a higher portion of the initial image than in CIFAR-10: we crop images of size 224x224 from images of size 256x256.

\section{Effect of number of Monte Carlo samples on ImageNet}
\label{app:monte carlo samples}

\begin{figure}[!t]
    \centering
    \includegraphics[width=\textwidth]{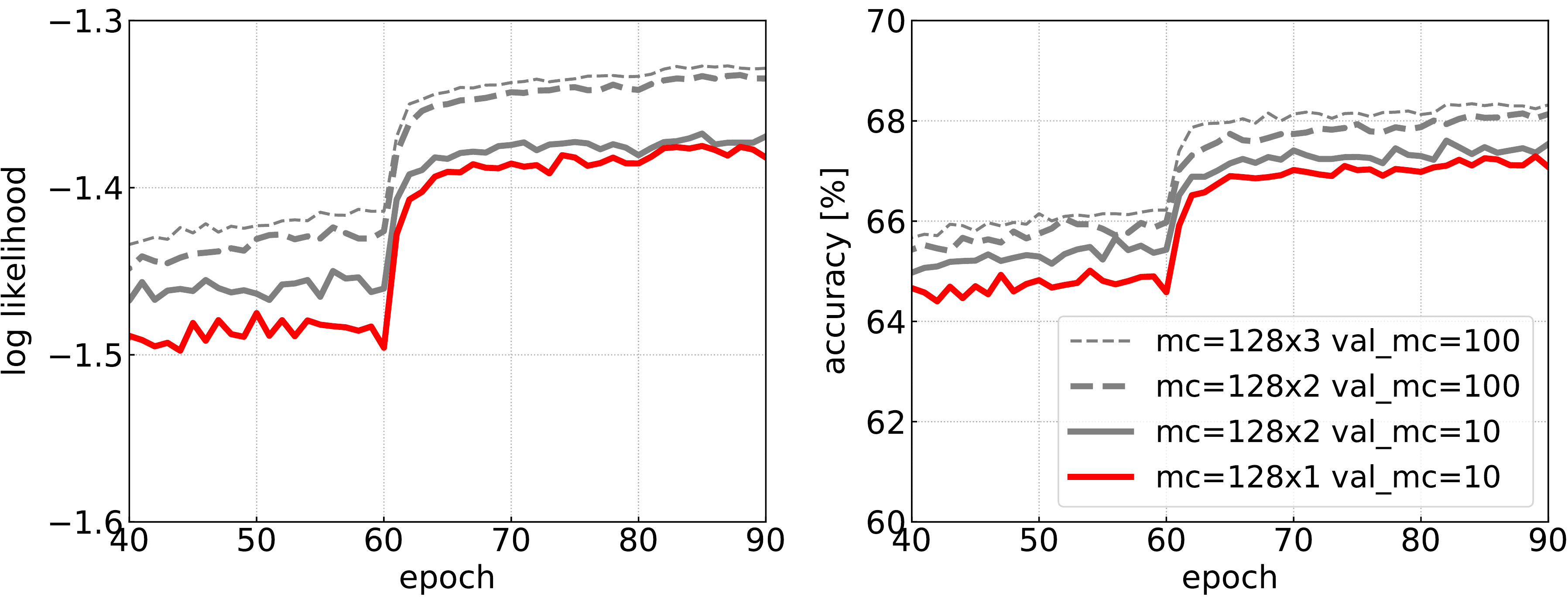}
    \includegraphics[width=\textwidth]{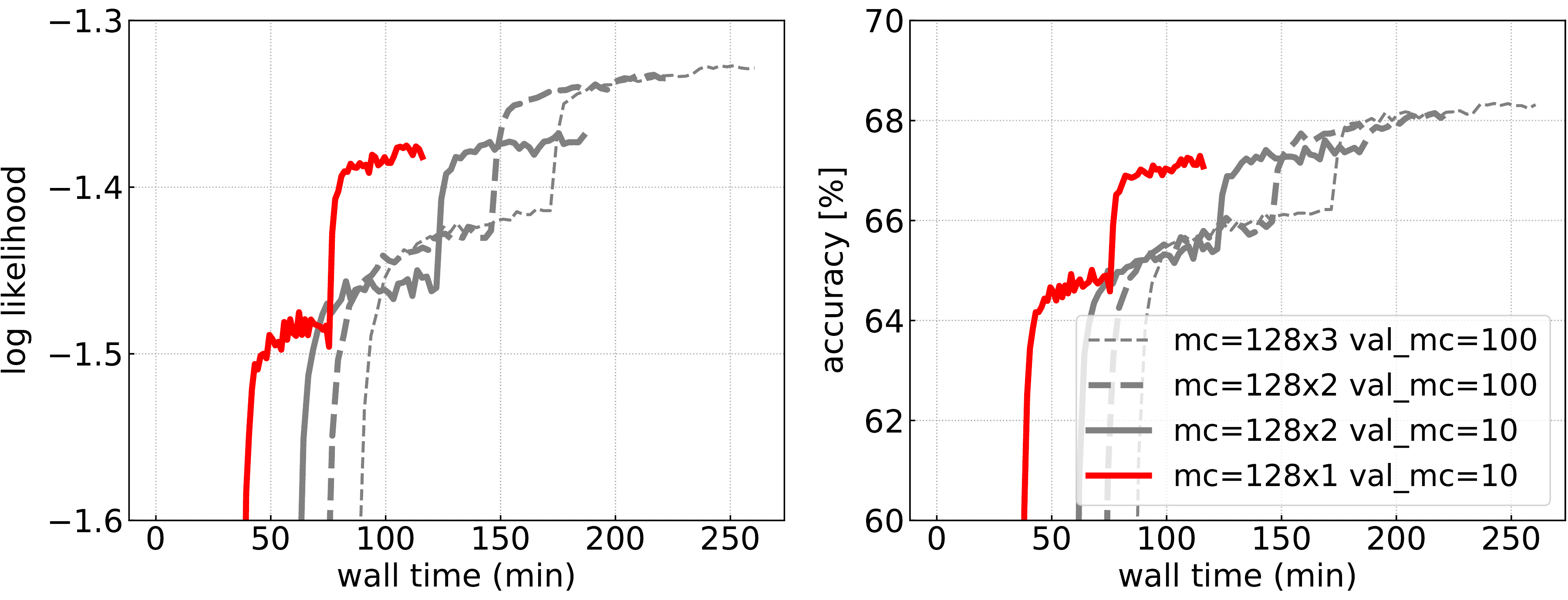}
    \caption{Effect of number of training and testing Monte Carlo samples on validation accuracy and log loss for VOGN on ResNet-18 on ImageNet.}
    \label{fig:resnet18 imagenet vogn mc samples accuracy}
\end{figure}

In the paper, we report results for training ResNet-18 on ImageNet using 128 GPUs, with 1 independent Monte-Carlo (MC) sample per process during training (\texttt{mc=128x1}), and 10 MC samples per validation image (\texttt{val\_mc$=10$}). We now show that increasing either of training or testing MC samples improves performance (validation accuracy and log likelihood) at the cost of increased computation time. See Figure \ref{fig:resnet18 imagenet vogn mc samples accuracy}.

Increasing the number of training MC samples per process reduces noise during training. This effect is observed when training on CIFAR-10, where multiple MC samples per process are required to stabilise training. On ImageNet, we have much larger minibatch size (4,096 instead of 256) and more parallel processes (128 not 8), and so training with 1 MC sample per process is still stable. However, as shown in Figure \ref{fig:resnet18 imagenet vogn mc samples accuracy}, increasing the number of training MC samples per process to from 1 to 2 speeds up convergence per epoch, and reaches a better converged solution. The time per epoch (and hence total runtime) also increases by approximately a factor of 1.5. Increasing the number of train MC samples per process to 3 does not increase final test performance significantly.

Increasing the number of testing MC samples from 10 to 100 (on the same trained model) also results in better generalisation: the train accuracy and log likelihood are unchanged, but the validation accuracy and log likelihood increase. However, as we run an entire validation on each epoch, increasing validation MC samples also increases run-time.

These results show that, if more compute is available to the user, they can improve VOGN's performance by improving the MC approximation at either (or both) train-time or test-time (up to a limit).

\section{MC-dropout's sensitivity to dropout rate}
\label{app:MC Dropout sensitivity}

\begin{figure}[H]
    \centering
    \includegraphics[width=1\textwidth]{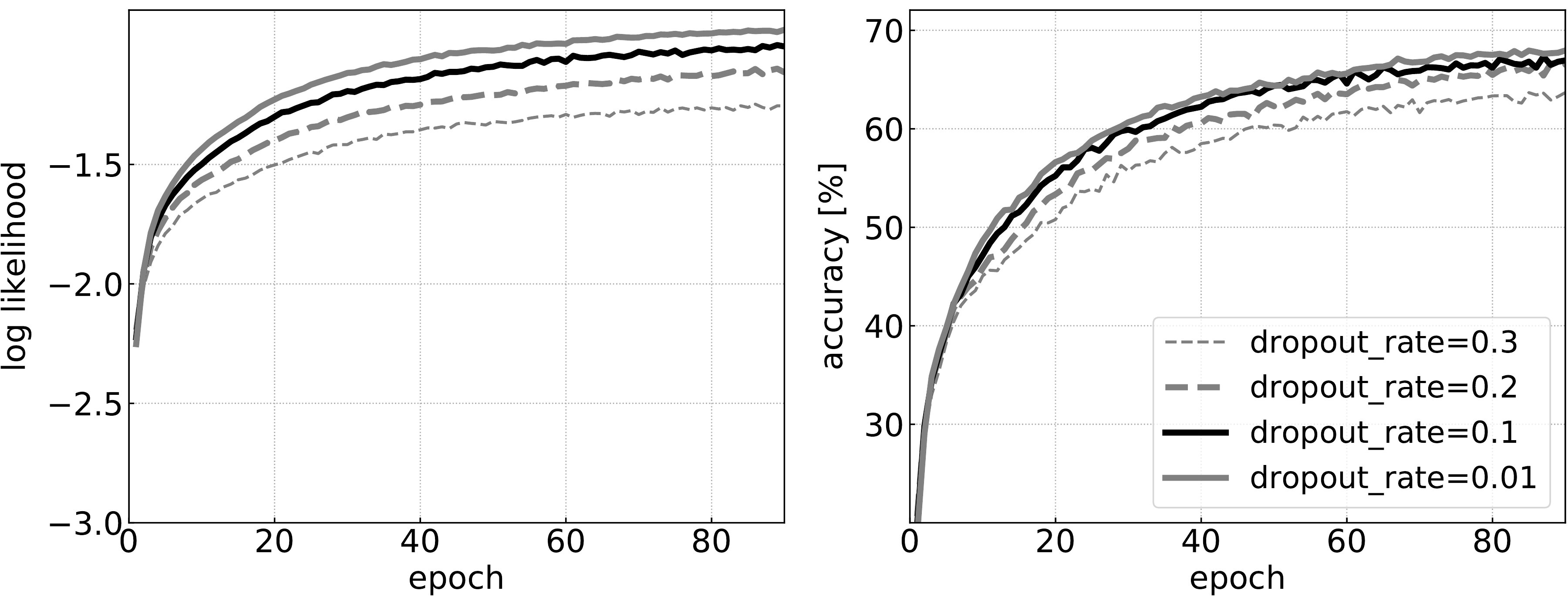}
        \caption{Effect of changing the dropout rate in MC-dropout, training LeNet-5 on CIFAR-10. When $p=0.01$, the train-test gap on accuracy and log likelihood is very high (10.3\% and 0.34 respectively). When $p=0.1$, gaps are 1.4\% and 0.04 respectively. When $p=0.2$, the gaps are -7.71\% and -0.02 respectively. We therefore choose $p=0.1$ as it has high accuracy and log likelihood, and small train-test gap.}
    \label{fig:cifar10_lenet5_mcdropout_sensitivity}
\end{figure}

We show MC-dropout's sensitivity to dropout rate, $p$, in this Appendix. We tune MC-dropout as best as we can, finding that $p=0.1$ works best for all architectures trained on CIFAR-10 (see Figure \ref{fig:cifar10_lenet5_mcdropout_sensitivity} for the dropout rate's sensitivity on LeNet-5 as an example). On ResNet-18 trained on ImageNet, we find that MC-dropout is extremely sensitive to dropout rate, with even $p=0.1$ performing badly. We therefore use $p=0.05$ for MC-dropout experiments on ImageNet. This high sensitivity to dropout rate is an issue with MC-dropout as a method.

\begin{figure}[H]
    \centering
    \includegraphics[width=1\textwidth]{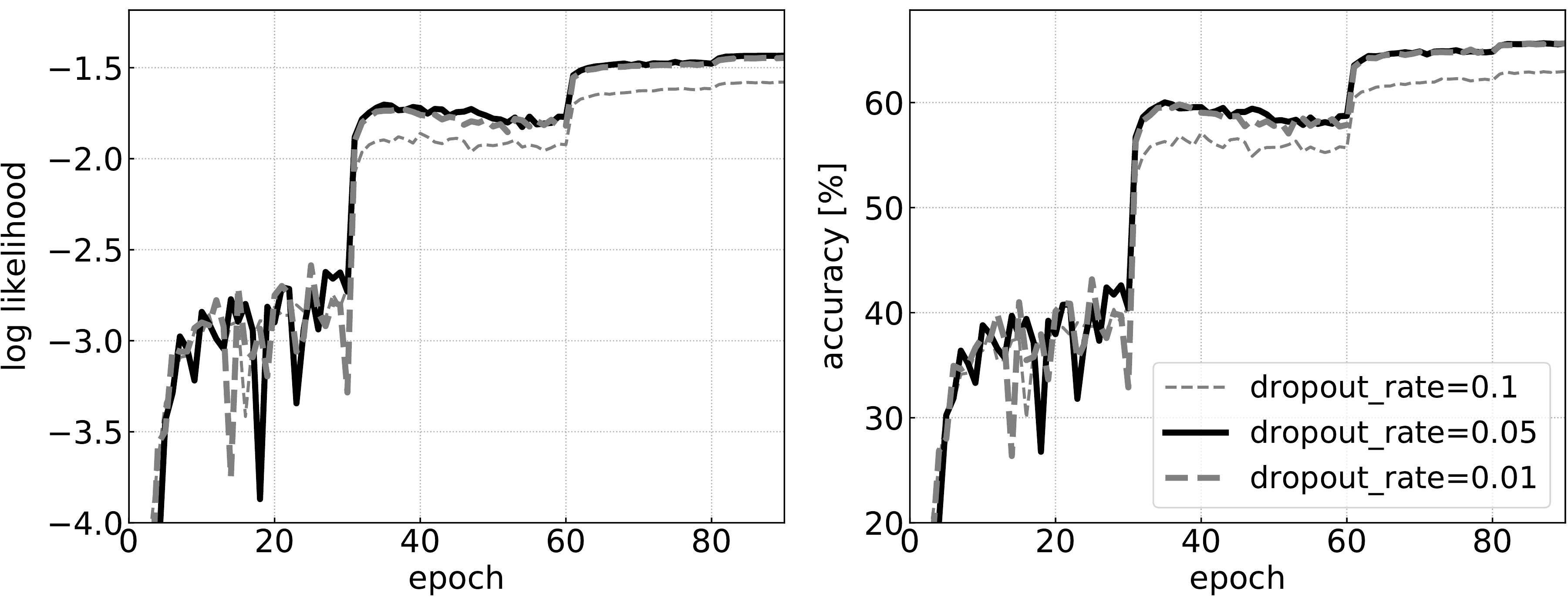}
    \caption{Effect of changing the dropout rate in MC-dropout, training Resnet-18 on ImageNet. We use $p=0.05$ for our results.}
    \label{fig:resnet18_imagenet_mcdropout_sensitivity}
\end{figure}

\section{Uncertainty metrics}
\label{app:uncertainty metrics}

We use several approaches to compare uncertainty estimates obtained by each optimiser. We follow the same methodology for all optimisers: first, tune hyperparameters to obtain good accuracy on the validation set. Then, test on uncertainty metrics. For multi-class classification problems, all of these are based on the predictive probabilities. For non-Bayesian approaches, we compute the probabilities for a validation input $\vx_i$ as $\hat{p}_{ik} := p(y_i=k|\vx_i, \vparam_*)$, where $\vparam_*$ is the weight vector of the DNN whose uncertainty we are estimating. For Bayesian methods, we can compute the predictive probabilities for each validation example $\vx_i$ as follows:
\begin{align*}
    \hat{p}_{ik} := \int p(y_i=k|\vx_i, \vparam) p(\vparam|\data) d\vparam
    \approx  \int p(y_i=k|\vx_i, \vparam) q(\vparam) d\vparam
    \approx  \frac{1}{C} \sum_{c=1}^C p(y_i=k|\vx_i, \vparam^{(c)} ) ,
\end{align*}
where $\vparam^{(c)} \sim q(\vparam)$ are samples from the Gaussian approximation returned by a variational method. 
We use 10 MC samples at validation-time for VOGN and MC-dropout (the effect of changing number of validation MC samples is shown in Appendix \ref{app:monte carlo samples}). This increases the computational cost during testing for these methods when compared to Adam or SGD.

Using the estimates $\hat{p}_{ik}$, we use three methods to compare uncertainties: validation log loss, AUROC and calibration curves. We also compare uncertainty performance by looking at model outputs when exposed to out-of-distribution data.

\textbf{Validation log likelihood.} Log likelihood (or log loss) is a common uncertainty metric. We consider a validation set of $N_{Va}$ examples. For an input $\vx_i$, denote the true label by $\vy_i$, a 1-of-$K$ encoded vector with 1 at the true label and 0 elsewhere. Denote the full vector of all validation outputs by $\vy$.
Similarly, denote the vector of all probabilities $\hat{p}_{ik}$ by $\vp$, where $k \in \{ 1,...,K \}$. The validation log likelihood is defined as
$
    \loss(\vy, \hat{\vp}) := \frac{1}{N_{Va}} \sum_{i = 1}^{N_{Va}} \sum_{k=1}^K y_{ik} \log \hat{p}_{ik}
$.

Tables \ref{table:full results} and \ref{table:full results with std devs} show final validation (negative) log likelihood. 
VOGN performs very well on this metric (aside from on CIFAR-10/AlexNet, with or without DA, where MC-dropout performs the best).
All final validation log likelihoods are very similar, with VOGN usually performing similarly to the other best-performing optimisers (usually MC-dropout).

\textbf{Area Under ROC curves (AUROC).} We consider Receiver Operating Characteristic (ROC) curves for our multi-way classification tasks. A potential way that we may care about uncertainty measurements would be to discard uncertain examples by thresholding each validation input's predicted class' softmax output, marking them as too ambiguous to belong to a class. We can then consider the remaining validation inputs to either be correctly or incorrectly classified, and calculate the True Positive Rate (TPR) and False Positive Rate (FPR) accordingly. The ROC curve is summarised by its Area Under Curve (AUROC), reported in Table \ref{table:full results}. This metric is useful to compare uncertainty performance in conjunction with the other metrics we use. The AUROC results are very similar between optimisers, particularly on ImageNet, although MC-dropout performs marginally better than the others, including VOGN. On all but one CIFAR-10 experiment (AlexNet, without DA), VOGN performs the best, or tied best. Adam performs the worst, but is surprisingly good in CIFAR-10/ResNet-18.

\begin{figure}[ht]
    \centering
    \includegraphics[width=0.8\textwidth]{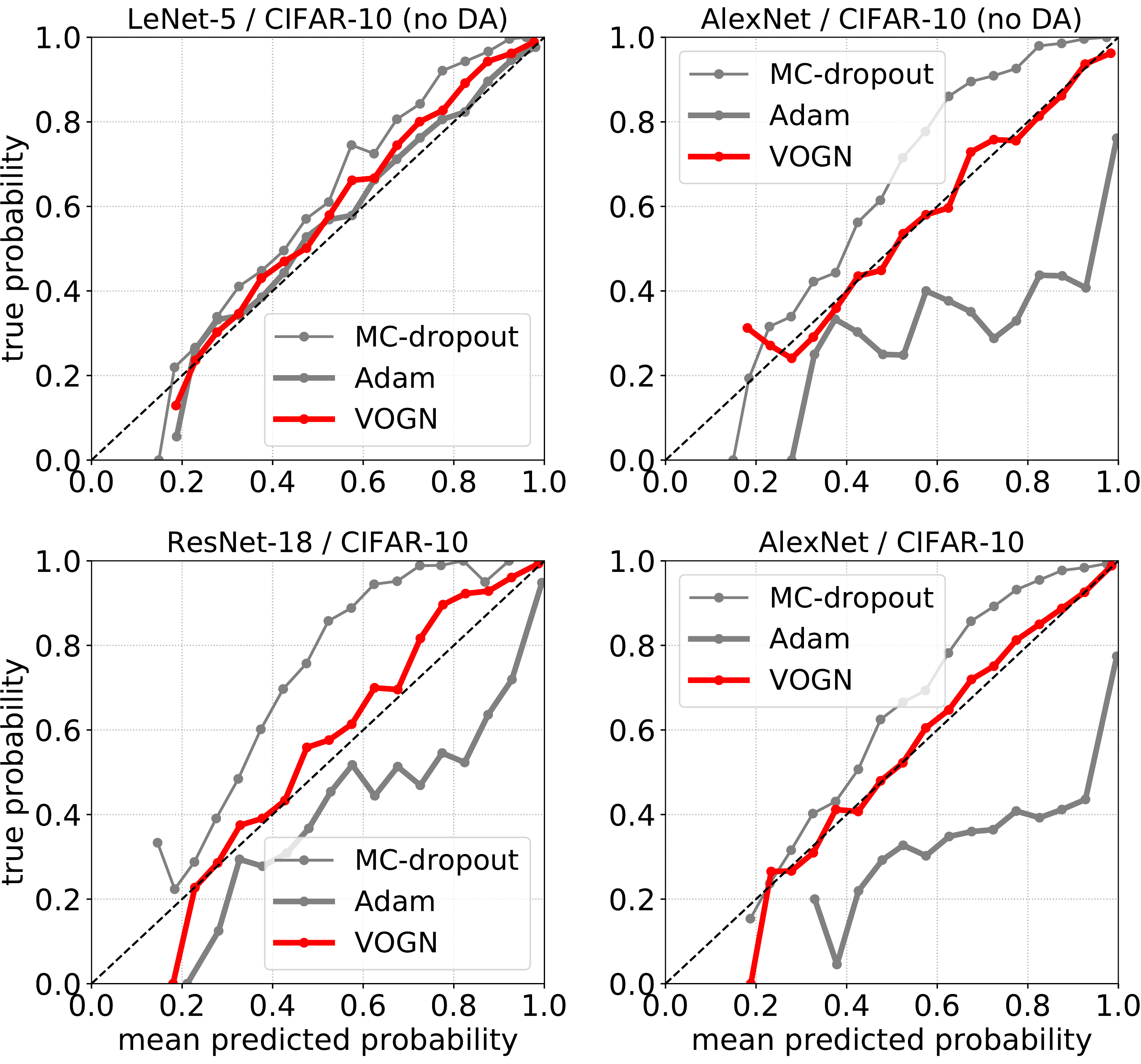}
    \caption{Calibration curves comparing VOGN, Adam and MC-dropout for final trained models trained on CIFAR-10. VOGN is extremely well-calibrated compared to the other two optimisers (except for LeNet-5, where all optimisers peform well). The calibration curve for ResNet-18 trained on ImageNet is in Figure \ref{fig:resnet18_imagenet_new}.}
    \label{fig:cc_cifar10}
\end{figure}

\textbf{Calibration Curves.} Calibration curves \citep{degroot1983comparison} test how well-calibrated a model is by plotting true accuracy as a function of the model's predicted accuracy $\hat{p}_{ik}$ (we only consider the predicted class' $\hat{p}_{ik}$). Perfectly calibrated models would follow the $y=x$ diagonal line on a calibration curve. We approximate this curve by binning the model's predictions into $M=20$ bins, as is often done.
We show calibration curves in Figures \ref{fig:resnet18_imagenet_new} and \ref{fig:cc_cifar10}.
We can also consider the \textbf{Expected Calibration Error (ECE)} metric \citep{naeini2015obtaining, guo2017oncalibration}, reported in Table \ref{table:full results}. ECE calculates the expected error between the true accuracy and the model's predicted accuracy, averaged over all validation examples, again approximated by using $M$ bins. 
Across all datasets and architectures, with the exception of LeNet-5 (which we have argued causes underfitting), VOGN usually has better calibration curves and better ECE than competing optimisers.
Adam is consistently over-confident, with the calibration curve below the diagonal. Conversely, MC-dropout is usually under-confident, with too much noise, as mentioned earlier. The exception to this is on ImageNet, where MC-dropout performs well: we excessively tuned the MC-dropout rate to achieve this (see Appendix \ref{app:MC Dropout sensitivity}).

\section{Out-of-distribution experimental setup and additional results}
\label{app:out-of-distribution results}

We use experiments from the out-of-distribution tests literature \citep{hendrycks2017baseline, lee2018training, devries2018learning, liang2018enhancing}, comparing VOGN to Adam and MC-dropout. Using trained architectures (LeNet-5, AlexNet and ResNet-18) on CIFAR-10, we test on SVHN, LSUN (crop) and LSUN (re-size) as out-of-distribution datasets, with the in-distribution data given by the validation set of CIFAR-10 (10,000 images). The entire training set of SVHN (73,257 examples, 10 classes) \citep{netzer2011reading} is used. The test set of LSUN (Large-scale Scene UNderstanding dataset \citep{yu2015lsun}, 10,000 images from 10 different scenes) is randomly cropped to obtain LSUN (crop), and is down-sampled to obtain LSUN (re-size). These out-of-distribution datasets have no similar classes to CIFAR-10.

\begin{figure}[htbp]
    \centering
    \includegraphics[width=\textwidth]
    {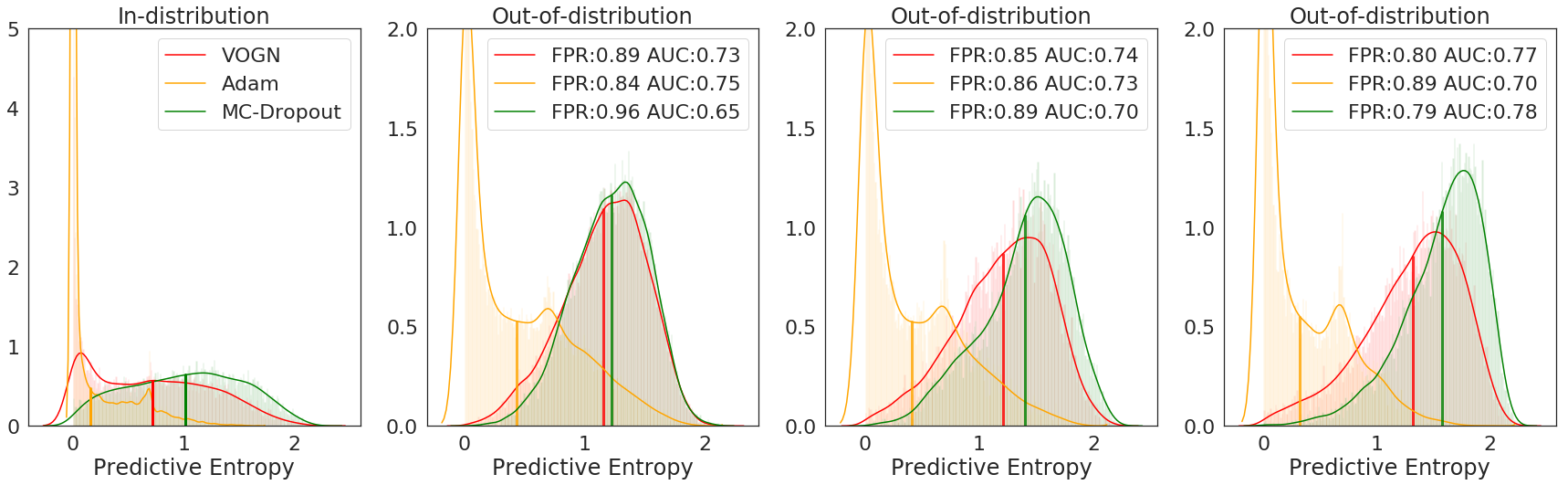}
    \caption{
    Histograms of predictive entropy for out-of-distribution tests for AlexNet trained on CIFAR-10 with data augmentation. Going from left to right, the inputs are: the in-distribution dataset (CIFAR-10), followed by out-of-distribution data: SVHN, LSUN (crop), LSUN (resize). Also shown are the AUROC metric (higher is better) and FPR at 95\% TPR metric (lower is better), averaged over 3 runs. The standard deviations are very small and so not reported here.
    }
    \label{fig:alexnet ood histograms}
\end{figure}

\begin{figure}[htbp]
    \centering
    \includegraphics[width=\textwidth]
    {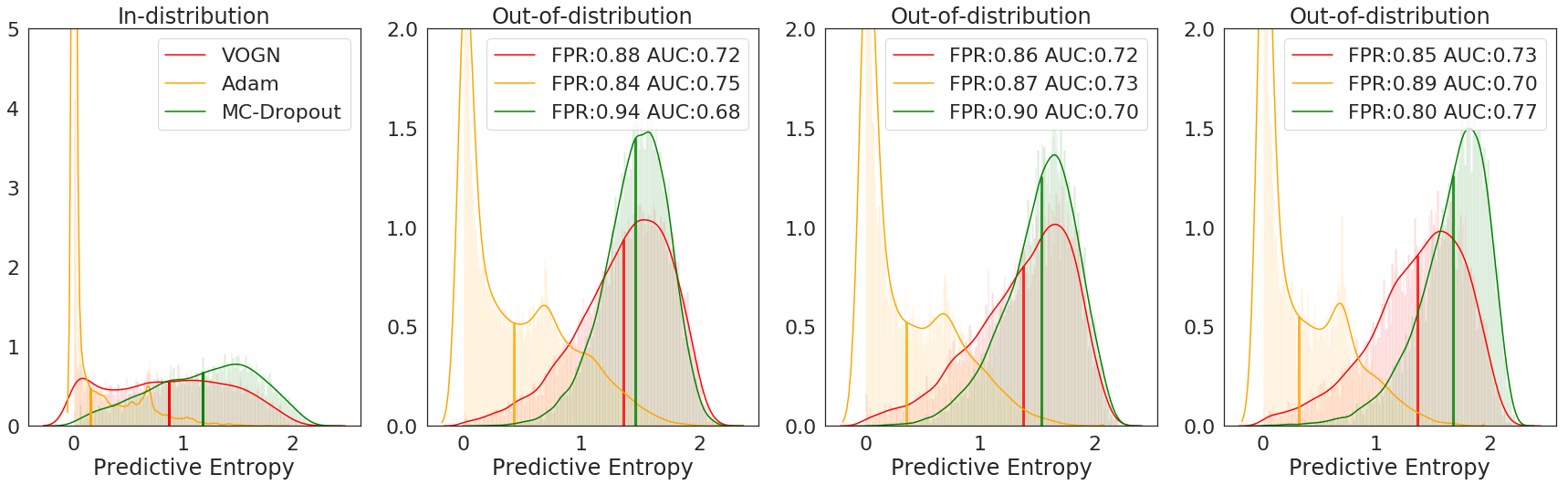}
    \caption{
    Histograms of predictive entropy for out-of-distribution tests for AlexNet trained on CIFAR-10 without data augmentation. Going from left to right, the inputs are: the in-distribution dataset (CIFAR-10), followed by out-of-distribution data: SVHN, LSUN (crop), LSUN (resize). Also shown are the AUROC metric (higher is better) and FPR at 95\% TPR metric (lower is better), averaged over 3 runs. The standard deviations are very small and so not reported here.
    }
    \label{fig:alexnet noda ood histograms}
\end{figure}

\begin{figure}[htbp]
    \centering
    \includegraphics[width=\textwidth]
    {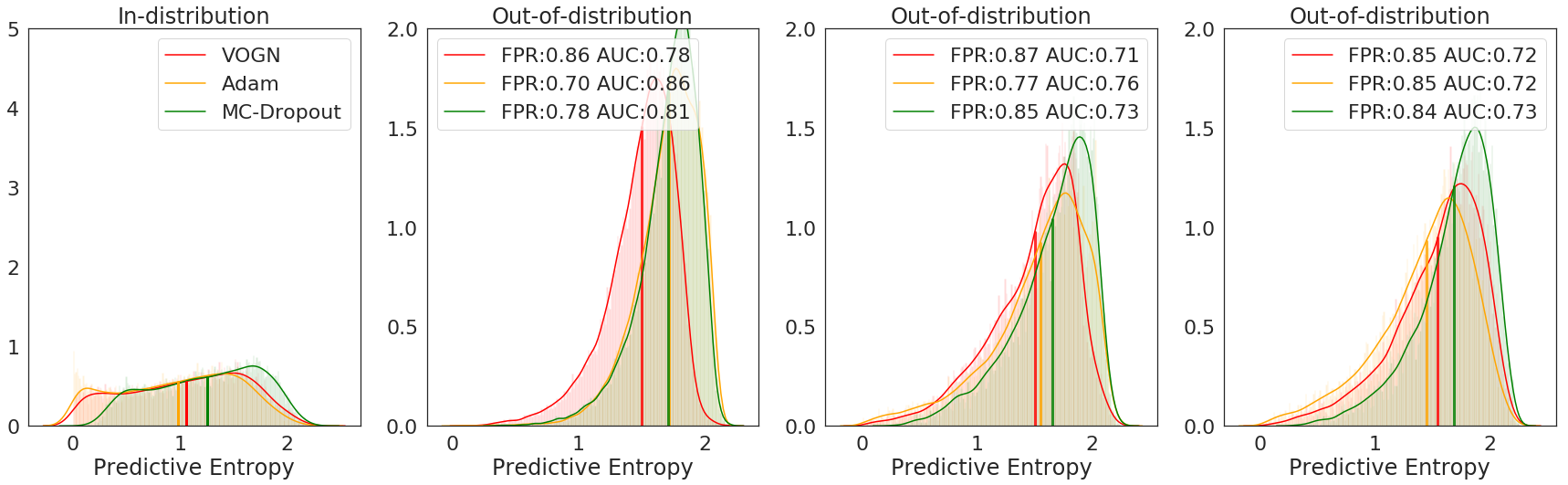}
    \caption{
    Histograms of predictive entropy for out-of-distribution tests for LeNet-5 trained on CIFAR-10 without data augmentation. Going from left to right, the inputs are: the in-distribution dataset (CIFAR-10), followed by out-of-distribution data: SVHN, LSUN (crop), LSUN (resize). Also shown are the AUROC metric (higher is better) and FPR at 95\% TPR metric (lower is better), averaged over 3 runs. The standard deviations are very small and so not reported here.
    }
    \label{fig:lenet5 ood histograms}
\end{figure}

Similar to the literature \citep{hendrycks2017baseline, lakshminarayanan2017simple}, we use 3 metrics to test performance on out-of-distribution data. Firstly, we plot histograms of predictive entropy for the in-distribution and out-of-distribution datasets, seen in Figure \ref{fig:resnet18 ood histograms}, \ref{fig:alexnet ood histograms}, \ref{fig:alexnet noda ood histograms} and \ref{fig:lenet5 ood histograms}. Predictive entropy is given by $\sum_{k=1}^K{-\hat{p}_{ik}\log{\hat{p}_{ik}}}$. Ideally, on out-of-distribution data, a model would have high predictive entropy, indicating it is unsure of which class the input image belongs to. In contrast, for in-distribution data, good models should have many examples with low entropy, as they should be confident of many input examples' (correct) class. We also compare AUROC and FPR at 95\% TPR, also reported in the figures. By thresholding the most likely class' softmax output, we assign high uncertainty images to belong to an unknown class. This allows us to calculate the FPR and TPR, allowing the ROC curve to be plotted, and the AUROC to be calculated.

We show results on AlexNet in Figure \ref{fig:alexnet ood histograms} and \ref{fig:alexnet noda ood histograms} (trained on CIFAR-10 with DA and without DA respectively) and on LeNet-5 in Figure \ref{fig:lenet5 ood histograms}. Results on ResNet-18 is in Figure \ref{fig:resnet18 ood histograms}. These results are discussed in Section \ref{sec:uncertainty}.

\section{Author contributions statement}
List of Authors: 
Kazuki Osawa, Siddharth Swaroop, Anirudh Jain, Runa Eschenhagen, Richard E. Turner, Rio Yokota, Mohammad Emtiyaz Khan.

M.E.K., A.J., and R.E. conceived the original idea. This was also discussed with R.Y. and K.O. and then with S.S. and R.T. Eventually, all authors discussed and agreed with the main focus and ideas of this paper.

The first proof-of-concept was done by A.J. using LeNet-5 on CIFAR-10. This was then extended by K.O. who wrote the main PyTorch implementation, including the distributed version. R.E. fixed multiple issues in the implementation, and also pointed out an important issue regarding data augmentation. S.S., A.J., K.O., and R.E. together fixed this issue. K.O. conducted most of the large experiments (shown in Fig. \ref{fig:resnet18_imagenet_new} and \ref{fig:cifar_many_architectures}). The results shown in Fig. \ref{fig:resnet18_cifar10_vogn} was done by both K.O. and A.J. The BBB implementation was written by S.S. 

The experiments in Section \ref{sec:uncertainty} were performed by A.J. and S.S. The main ideas behind the experiments were conceived by S.S., A.J., and M.E.K. with many helpful suggestions from R.T.
R.E. performed the permuted MNIST experiment using VOGN for the continual-learning experiments, and S.S. obtained the baseline results for the same.

The main text of the paper was written by M.E.K. and S.S. The section on experiments was first written by S.S. and subsequently improved by A.J., K.O., and M.E.K. R.T. helped edit the manuscript. R.E. also helped in writing parts of the paper.

M.E.K. led the project with a significant help from S.S.. Computing resources and access to the HPCI systems were provided by R.Y.

\section{Changes in the camera-ready version compared to the submitted version}
\begin{itemize}
    \item We added an additional experiment on a continual learning task to show the effectiveness of VOGN (Figure~\ref{fig:permuted mnist}).
    \item In our experiments, we were using a damping factor $\gamma$. This was unfortunately missed in the submitted version, and we have now added it in Section~\ref{sec:methods}.
    \item We modified the notation for Noisy K-FAC algorithm at Appendix~\ref{app:noisykfac}.
    \item We updated the description of our implementation of the Gauss-Newton approximation at Appendix~\ref{app:GN implementation}. Previous description had some missing parts and was a bit unclear.
    \item We added a description on a new method OGN which we were using to tune hyperparameters of VOGN.
    We have added its results in Table~\ref{table:full results} and Table~\ref{table:full results with std devs}.
    The method details are in Appendix~\ref{app:ogn}.
    \item We added a description on how to tune VOGN to get good performance.
    \item We listed all training curves (epoch/time vs accuracy), including K-FAC, Noisy K-FAC, and OGN, along with the corresponding calibration curves in Figure~\ref{fig:all_curves}.
\end{itemize}

\begin{thebibliography}{56}
\providecommand{\natexlab}[1]{#1}
\providecommand{\url}[1]{\texttt{#1}}
\expandafter\ifx\csname urlstyle\endcsname\relax
  \providecommand{\doi}[1]{doi: #1}\else
  \providecommand{\doi}{doi: \begingroup \urlstyle{rm}\Url}\fi

\bibitem[Anderson and Peterson(1987)]{anderson1987mean}
James~R Anderson and Carsten Peterson.
\newblock A mean field theory learning algorithm for neural networks.
\newblock \emph{Complex Systems}, 1:\penalty0 995--1019, 1987.

\bibitem[Barber and Bishop(1998)]{barber1998ensemble}
David Barber and Christopher~M Bishop.
\newblock Ensemble learning in {Bayesian} neural networks.
\newblock \emph{Generalization in Neural Networks and Machine Learning},
  168:\penalty0 215--238, 1998.

\bibitem[Bishop(2006)]{bishop2006pattern}
Christopher~M. Bishop.
\newblock \emph{Pattern Recognition and Machine Learning (Information Science
  and Statistics)}.
\newblock Springer-Verlag, Berlin, Heidelberg, 2006.
\newblock ISBN 0387310738.

\bibitem[Blundell et~al.(2015)Blundell, Cornebise, Kavukcuoglu, and
  Wierstra]{blundell2015weight}
Charles Blundell, Julien Cornebise, Koray Kavukcuoglu, and Daan Wierstra.
\newblock Weight uncertainty in neural networks.
\newblock In \emph{International Conference on Machine Learning}, pages
  1613--1622, 2015.

\bibitem[Bottou et~al.(2016)Bottou, Curtis, and
  Nocedal]{bottou2016optimization}
L{\'e}on Bottou, Frank~E Curtis, and Jorge Nocedal.
\newblock Optimization methods for large-scale machine learning.
\newblock \emph{arXiv preprint arXiv:1606.04838}, 2016.

\bibitem[Bradshaw et~al.(2017)Bradshaw, Matthews, and
  Ghahramani]{bradshaw2017adversarial}
John Bradshaw, Alexander G de~G Matthews, and Zoubin Ghahramani.
\newblock Adversarial examples, uncertainty, and transfer testing robustness in
  {G}aussian process hybrid deep networks.
\newblock \emph{arXiv preprint arXiv:1707.02476}, 2017.

\bibitem[DeGroot and Fienberg(1983)]{degroot1983comparison}
Morris~H. DeGroot and Stephen~E. Fienberg.
\newblock The comparison and evaluation of forecasters.
\newblock \emph{The Statistician: Journal of the Institute of Statisticians},
  32:\penalty0 12--22, 1983.

\bibitem[DeVries and Taylor(2018)]{devries2018learning}
Terrance DeVries and Graham~W. Taylor.
\newblock Learning confidence for out-of-distribution detection in neural
  networks.
\newblock \emph{arXiv preprint arXiv:1802.04865}, 2018.

\bibitem[Gal and Ghahramani(2016)]{yarin16dropout}
Yarin Gal and Zoubin Ghahramani.
\newblock Dropout as a {Bayesian} approximation: Representing model uncertainty
  in deep learning.
\newblock In \emph{International Conference on Machine Learning}, pages
  1050--1059, 2016.

\bibitem[Ghosal and Van~der Vaart(2017)]{ghosal2017fundamentals}
S.~Ghosal and A.~Van~der Vaart.
\newblock \emph{Fundamentals of nonparametric {B}ayesian inference}, volume~44.
\newblock Cambridge {U}niversity {P}ress, 2017.

\bibitem[Glorot and Bengio(2010)]{glorot2010understanding}
Xavier Glorot and Yoshua Bengio.
\newblock Understanding the difficulty of training deep feedforward neural
  networks.
\newblock In \emph{Proceedings of the thirteenth international conference on
  artificial intelligence and statistics}, pages 249--256, 2010.

\bibitem[Goodfellow(2015)]{goodfellow2015efficient}
Ian Goodfellow.
\newblock {Efficient Per-Example Gradient Computations}.
\newblock \emph{ArXiv e-prints}, October 2015.

\bibitem[Goyal et~al.(2017)Goyal, Doll{\'{a}}r, Girshick, Noordhuis,
  Wesolowski, Kyrola, Tulloch, Jia, and He]{goyal2017accurate}
Priya Goyal, Piotr Doll{\'{a}}r, Ross~B. Girshick, Pieter Noordhuis, Lukasz
  Wesolowski, Aapo Kyrola, Andrew Tulloch, Yangqing Jia, and Kaiming He.
\newblock Accurate, large minibatch {SGD:} training imagenet in 1 hour.
\newblock \emph{CoRR}, abs/1706.02677, 2017.

\bibitem[Graves(2011)]{graves2011practical}
Alex Graves.
\newblock Practical variational inference for neural networks.
\newblock In \emph{Advances in Neural Information Processing Systems}, pages
  2348--2356, 2011.

\bibitem[Guo et~al.(2017)Guo, Pleiss, Sun, and
  Weinberger]{guo2017oncalibration}
Chuan Guo, Geoff Pleiss, Yu~Sun, and Kilian~Q Weinberger.
\newblock On calibration of modern neural networks.
\newblock In \emph{Proceedings of the 34th International Conference on Machine
  Learning-Volume 70}, pages 1321--1330. JMLR. org, 2017.

\bibitem[Hendrycks and Gimpel(2017)]{hendrycks2017baseline}
Dan Hendrycks and Kevin Gimpel.
\newblock A baseline for detecting misclassified and out-of-distribution
  examples in neural networks.
\newblock In \emph{International Conference on Learning Representations}, 2017.

\bibitem[Hinton et~al.(2012)Hinton, Deng, Yu, Dahl, Mohamed, Jaitly, Senior,
  Vanhoucke, Nguyen, Kingsbury, et~al.]{hinton2012deep}
Geoffrey Hinton, Li~Deng, Dong Yu, George Dahl, Abdel-rahman Mohamed, Navdeep
  Jaitly, Andrew Senior, Vincent Vanhoucke, Patrick Nguyen, Brian Kingsbury,
  et~al.
\newblock Deep neural networks for acoustic modeling in speech recognition.
\newblock \emph{IEEE Signal processing magazine}, 29, 2012.

\bibitem[Hinton and Van~Camp(1993)]{hinton1993keeping}
Geoffrey~E Hinton and Drew Van~Camp.
\newblock Keeping the neural networks simple by minimizing the description
  length of the weights.
\newblock In \emph{Annual Conference on Computational Learning Theory}, pages
  5--13, 1993.

\bibitem[Hoeting et~al.(1999)Hoeting, Madigan, Raftery, and
  Volinsky]{hoeting1999bayesian}
Jennifer~A Hoeting, David Madigan, Adrian~E Raftery, and Chris~T Volinsky.
\newblock Bayesian model averaging: a tutorial.
\newblock \emph{Statistical science}, pages 382--401, 1999.

\bibitem[Ioffe and Szegedy(2015)]{ioffe2015batchnorm}
Sergey Ioffe and Christian Szegedy.
\newblock Batch normalization: Accelerating deep network training by reducing
  internal covariate shift.
\newblock \emph{CoRR}, abs/1502.03167, 2015.
\newblock URL \url{http://arxiv.org/abs/1502.03167}.

\bibitem[Khan(2012)]{khan2012variational}
Mohammad Khan.
\newblock \emph{Variational learning for latent Gaussian model of discrete
  data}.
\newblock PhD thesis, University of British Columbia, 2012.

\bibitem[Khan and Lin(2017)]{khan2017conjugate}
Mohammad~Emtiyaz Khan and Wu~Lin.
\newblock Conjugate-computation variational inference: converting variational
  inference in non-conjugate models to inferences in conjugate models.
\newblock In \emph{International Conference on Artificial Intelligence and
  Statistics}, pages 878--887, 2017.

\bibitem[Khan and Nielsen(2018)]{khan2018fast1}
Mohammad~Emtiyaz Khan and Didrik Nielsen.
\newblock Fast yet simple natural-gradient descent for variational inference in
  complex models.
\newblock In \emph{2018 International Symposium on Information Theory and Its
  Applications (ISITA)}, pages 31--35. IEEE, 2018.

\bibitem[Khan et~al.(2018)Khan, Nielsen, Tangkaratt, Lin, Gal, and
  Srivastava]{khan2018fast}
Mohammad~Emtiyaz Khan, Didrik Nielsen, Voot Tangkaratt, Wu~Lin, Yarin Gal, and
  Akash Srivastava.
\newblock Fast and scalable {B}ayesian deep learning by weight-perturbation in
  {A}dam.
\newblock In \emph{International Conference on Machine Learning}, pages
  2616--2625, 2018.

\bibitem[Kingma and Ba(2015)]{kingma2014adam}
Diederik Kingma and Jimmy Ba.
\newblock Adam: A method for stochastic optimization.
\newblock In \emph{International Conference on Learning Representations}, 2015.

\bibitem[Kingma et~al.(2015)Kingma, Salimans, and
  Welling]{kingma2015variational}
Diederik~P Kingma, Tim Salimans, and Max Welling.
\newblock Variational dropout and the local reparameterization trick.
\newblock In \emph{Advances in Neural Information Processing Systems}, pages
  2575--2583, 2015.

\bibitem[Kirkpatrick et~al.(2017)Kirkpatrick, Pascanu, Rabinowitz, Veness,
  Desjardins, Rusu, Milan, Quan, Ramalho, Grabska-Barwinska,
  et~al.]{kirkpatrick2017overcoming}
James Kirkpatrick, Razvan Pascanu, Neil Rabinowitz, Joel Veness, Guillaume
  Desjardins, Andrei~A Rusu, Kieran Milan, John Quan, Tiago Ramalho, Agnieszka
  Grabska-Barwinska, et~al.
\newblock Overcoming catastrophic forgetting in neural networks.
\newblock \emph{Proceedings of the national academy of sciences}, 114\penalty0
  (13):\penalty0 3521--3526, 2017.

\bibitem[Krizhevsky and Hinton(2009)]{krizhevsky2009cifar}
Alex Krizhevsky and Geoffrey Hinton.
\newblock Learning multiple layers of features from tiny images.
\newblock Technical report, Citeseer, 2009.

\bibitem[Krizhevsky et~al.(2012)Krizhevsky, Sutskever, and
  Hinton]{krizhevsky2012imagenet}
Alex Krizhevsky, Ilya Sutskever, and Geoffrey~E Hinton.
\newblock Imagenet classification with deep convolutional neural networks.
\newblock In \emph{Advances in neural information processing systems}, pages
  1097--1105, 2012.

\bibitem[Lakshminarayanan et~al.(2017)Lakshminarayanan, Pritzel, and
  Blundell]{lakshminarayanan2017simple}
Balaji Lakshminarayanan, Alexander Pritzel, and Charles Blundell.
\newblock Simple and scalable predictive uncertainty estimation using deep
  ensembles.
\newblock In \emph{Advances in Neural Information Processing Systems 30}, pages
  6402--6413. Curran Associates, Inc., 2017.

\bibitem[Lee et~al.(2018)Lee, Lee, Lee, and Shin]{lee2018training}
Kimin Lee, Honglak Lee, Kibok Lee, and Jinwoo Shin.
\newblock Training confidence-calibrated classifiers for detecting
  out-of-distribution samples.
\newblock In \emph{International Conference on Learning Representations}, 2018.

\bibitem[Liang et~al.(2018)Liang, Li, and Srikant]{liang2018enhancing}
Shiyu Liang, Yixuan Li, and R.~Srikant.
\newblock Enhancing the reliability of out-of-distribution image detection in
  neural networks.
\newblock In \emph{International Conference on Learning Representations}, 2018.

\bibitem[Lopez-Paz and Ranzato(2017)]{lopez2017gradient}
David Lopez-Paz and Marc~Aurelio Ranzato.
\newblock Gradient episodic memory for continual learning.
\newblock In \emph{NIPS}, 2017.

\bibitem[Loshchilov and Hutter(2019)]{loshchilov2018decoupled}
Ilya Loshchilov and Frank Hutter.
\newblock Decoupled weight decay regularization.
\newblock In \emph{International Conference on Learning Representations}, 2019.

\bibitem[Mackay(1991)]{mackay1991thesis}
David Mackay.
\newblock \emph{Bayesian Methods for Adaptive Models}.
\newblock PhD thesis, California Institute of Technology, 1991.

\bibitem[MacKay(2003)]{mackay2003information}
David~JC MacKay.
\newblock \emph{Information theory, inference and learning algorithms}.
\newblock Cambridge university press, 2003.

\bibitem[Maddox et~al.(2019)Maddox, Garipov, Izmailov, Vetrov, and
  Wilson]{maddox2019simple}
Wesley Maddox, Timur Garipov, Pavel Izmailov, Dmitry Vetrov, and Andrew~Gordon
  Wilson.
\newblock A simple baseline for {B}ayesian uncertainty in deep learning.
\newblock \emph{arXiv preprint arXiv:1902.02476}, 2019.

\bibitem[Mandt et~al.(2017)Mandt, Hoffman, and Blei]{mandt2017stochastic}
Stephan Mandt, Matthew~D Hoffman, and David~M Blei.
\newblock Stochastic gradient descent as approximate {B}ayesian inference.
\newblock \emph{Journal of Machine Learning Research}, 18:\penalty0 1--35,
  2017.

\bibitem[Mikolov et~al.(2013)Mikolov, Chen, Corrado, and
  Dean]{mikolov2013efficient}
Tomas Mikolov, Kai Chen, Greg Corrado, and Jeffrey Dean.
\newblock Efficient estimation of word representations in vector space.
\newblock \emph{arXiv preprint arXiv:1301.3781}, 2013.

\bibitem[Naeini et~al.(2015)Naeini, Cooper, and
  Hauskrecht]{naeini2015obtaining}
Mahdi~Pakdaman Naeini, Gregory~F. Cooper, and Milos Hauskrecht.
\newblock Obtaining well calibrated probabilities using {B}ayesian binning.
\newblock In \emph{Proceedings of the Twenty-Ninth AAAI Conference on
  Artificial Intelligence}, AAAI'15, pages 2901--2907. AAAI Press, 2015.

\bibitem[Neal(1995)]{neal95}
Redford~M Neal.
\newblock \emph{{B}ayesian learning for neural networks}.
\newblock PhD thesis, University of Toronto, 1995.

\bibitem[Netzer et~al.(2011)Netzer, Wang, Coates, Bissacco, Wu, and
  Ng]{netzer2011reading}
Yuval Netzer, Tao Wang, Adam Coates, Alessandro Bissacco, Bo~Wu, and Andrew~Y.
  Ng.
\newblock Reading digits in natural images with unsupervised feature learning.
\newblock In \emph{NIPS Workshop on Deep Learning and Unsupervised Feature
  Learning 2011}, 2011.

\bibitem[Nguyen et~al.(2017)Nguyen, Li, Bui, and Turner]{nguyen2017variational}
Cuong~V Nguyen, Yingzhen Li, Thang~D Bui, and Richard~E Turner.
\newblock Variational continual learning.
\newblock \emph{arXiv preprint arXiv:1710.10628}, 2017.

\bibitem[Osawa et~al.(2018)Osawa, Tsuji, Ueno, Naruse, Yokota, and
  Matsuoka]{osawa2018secondorder}
Kazuki Osawa, Yohei Tsuji, Yuichiro Ueno, Akira Naruse, Rio Yokota, and Satoshi
  Matsuoka.
\newblock Second-order optimization method for large mini-batch: Training
  resnet-50 on imagenet in 35 epochs.
\newblock \emph{CoRR}, abs/1811.12019, 2018.

\bibitem[Riquelme et~al.(2018)Riquelme, Tucker, and Snoek]{riquelme2018deep}
Carlos Riquelme, George Tucker, and Jasper Snoek.
\newblock Deep {B}ayesian bandits showdown: An empirical comparison of
  {B}ayesian deep networks for {T}hompson sampling.
\newblock \emph{arXiv preprint arXiv:1802.09127}, 2018.

\bibitem[Ritter et~al.(2018)Ritter, Botev, and Barber]{ritter2018scalable}
Hippolyt Ritter, Aleksandar Botev, and David Barber.
\newblock A scalable {L}aplace approximation for neural networks.
\newblock In \emph{International Conference on Learning Representations}, 2018.

\bibitem[Russakovsky et~al.(2015)Russakovsky, Deng, Su, Krause, Satheesh, Ma,
  Huang, Karpathy, Khosla, Bernstein, et~al.]{russakovsky2015imagenet}
Olga Russakovsky, Jia Deng, Hao Su, Jonathan Krause, Sanjeev Satheesh, Sean Ma,
  Zhiheng Huang, Andrej Karpathy, Aditya Khosla, Michael Bernstein, et~al.
\newblock Imagenet large scale visual recognition challenge.
\newblock \emph{International journal of computer vision}, 115\penalty0
  (3):\penalty0 211--252, 2015.

\bibitem[Rusu et~al.(2016)Rusu, Rabinowitz, Desjardins, Soyer, Kirkpatrick,
  Kavukcuoglu, Pascanu, and Hadsell]{rusu2016progressive}
Andrei~A. Rusu, Neil~C. Rabinowitz, Guillaume Desjardins, Hubert Soyer, James
  Kirkpatrick, Koray Kavukcuoglu, Razvan Pascanu, and Raia Hadsell.
\newblock Progressive neural networks.
\newblock \emph{arXiv preprint arXiv:1606.04671}, 2016.

\bibitem[Saul et~al.(1996)Saul, Jaakkola, and Jordan]{saul1996mean}
Lawrence~K Saul, Tommi Jaakkola, and Michael~I Jordan.
\newblock Mean field theory for sigmoid belief networks.
\newblock \emph{Journal of Artificial Intelligence Research}, 4:\penalty0
  61--76, 1996.

\bibitem[Schwarz et~al.(2018)Schwarz, Luketina, Czarnecki, Grabska-Barwinska,
  Teh, Pascanu, and Hadsell]{schwarz2018progress}
Jonathan Schwarz, Jelena Luketina, Wojciech~M. Czarnecki, Agnieszka
  Grabska-Barwinska, Yee~Whye Teh, Razvan Pascanu, and Raia Hadsell.
\newblock Progress \& compress: A scalable framework for continual learning.
\newblock In \emph{International Conference on Machine Learning}, 2018.

\bibitem[Sutskever et~al.(2013)Sutskever, Martens, Dahl, and
  Hinton]{sutskever2013importance}
Ilya Sutskever, James Martens, George Dahl, and Geoffrey Hinton.
\newblock On the importance of initialization and momentum in deep learning.
\newblock In \emph{International conference on machine learning}, pages
  1139--1147, 2013.

\bibitem[Swaroop et~al.(2019)Swaroop, Nguyen, Bui, and
  Turner]{swaroop2019improving}
Siddharth Swaroop, Cuong~V. Nguyen, Thang~D. Bui, and Richard~E. Turner.
\newblock Improving and understanding variational continual learning.
\newblock \emph{arXiv preprint arXiv:1905.02099}, 2019.

\bibitem[Tieleman and Hinton(2012)]{hintonTieleman}
Tijmen Tieleman and Geoffrey Hinton.
\newblock {Lecture 6.5-{R}MSprop: Divide the gradient by a running average of
  its recent magnitude.}
\newblock \emph{COURSERA: Neural Networks for Machine Learning 4}, 2012.

\bibitem[Vovk(1990)]{Vovk:1990:AS:92571.92672}
V.~G. Vovk.
\newblock Aggregating strategies.
\newblock In \emph{Proceedings of the Third Annual Workshop on Computational
  Learning Theory}, COLT '90, pages 371--386, San Francisco, CA, USA, 1990.
  Morgan Kaufmann Publishers Inc.
\newblock ISBN 1-55860-146-5.

\bibitem[Yu et~al.(2015)Yu, Zhang, Song, Seff, and Xiao]{yu2015lsun}
Fisher Yu, Yinda Zhang, Shuran Song, Ari Seff, and Jianxiong Xiao.
\newblock {LSUN:} construction of a large-scale image dataset using deep
  learning with humans in the loop.
\newblock \emph{CoRR}, abs/1506.03365, 2015.

\bibitem[Zhang et~al.(2018)Zhang, Sun, Duvenaud, and Grosse]{zhang2018noisy}
Guodong Zhang, Shengyang Sun, David~K. Duvenaud, and Roger~B. Grosse.
\newblock Noisy natural gradient as variational inference.
\newblock \emph{arXiv preprint arXiv:1712.02390}, 2018.

\end{thebibliography}
\end{document}